\documentclass[10pt,journal,compsoc]{IEEEtran}

\ifCLASSOPTIONcompsoc
  \usepackage[nocompress]{cite}
\else
  \usepackage{cite}
\fi

\ifCLASSINFOpdf
\else
\fi
\usepackage[dvipsnames,table,xcdraw]{xcolor}
\usepackage[pagebackref=true,breaklinks=true,colorlinks,citecolor=ForestGreen]{hyperref}
\usepackage{ragged2e}
\usepackage{amssymb}
\usepackage{makecell}
\usepackage{multirow}
\usepackage{rotating}
\usepackage{array}
\usepackage{times}
\usepackage{graphicx}
\usepackage{psfrag}
\usepackage{subfigure}
\usepackage{enumitem}
\usepackage{url}
\usepackage{amsmath}
\usepackage{booktabs}
\usepackage{lscape}
\usepackage{verbatim}
\usepackage{overpic}
\usepackage{bbding}
\usepackage{bm}
\usepackage{pifont}

\usepackage{tabularx}
\usepackage{booktabs}
\usepackage[utf8]{inputenc}
\usepackage[T1]{fontenc}

\usepackage{amsthm}

\usepackage{tablefootnote}

\usepackage{array}
\usepackage{makecell}

\usepackage{orcidlink} 

\usepackage{algpseudocode}
\usepackage[ruled,linesnumbered]{algorithm2e}
\SetKwComment{Comment}{/* }{ */}

\usepackage{xr-hyper}
\usepackage[misc]{ifsym}

\usepackage{enumitem}
\usepackage{soul}

\begin{document}
%

\title{R-PGA: Robust Physical Adversarial Camouflage Generation via Relightable 3D Gaussian Splatting}

\author{
    Tianrui Lou$^{\orcidlink{0000-0002-5325-3445}}$, Siyuan Liang$^{\orcidlink{0000-0002-6154-0233}}$, Jiawei Liang$^{\orcidlink{0000-0003-1143-6873}}$,  Yuze Gao$^{\orcidlink{0009-0001-4409-7102}}$,  and Xiaochun Cao$^{\orcidlink{0000-0001-7141-708X}}$, ~\textit{Senior Member, IEEE}
\thanks{
Tianrui Lou, Jiawei Liang and Xiaochun Cao (Corresponding) are with School of Cyber Science and Technology, Shenzhen Campus, Sun Yat-sen University, Shenzhen 518107, China (e-mail: loutianrui@gmail.com, liangjw57@mail2.sysu.edu.cn, caoxiaochun@mail.sysu.edu.cn)}
\thanks{Siyuan Liang is with the School of Computing, National University of Singapore, 117417, Singapore.(e-mail: pandaliang521@gmail.com)}
\thanks{Yuze Gao is with the School of Intelligent Systems Engineering, Shenzhen Campus, Sun Yat-sen University, Shenzhen 518107, China.(e-mail: gaoyuze2023@gmail.com)}
}

\markboth{Submitted to IEEE Transactions on Pattern Analysis and Machine Intelligence}%
{Shell \MakeLowercase{\textit{et al.}}: Bare Demo of IEEEtran.cls for Computer Society Journals}

\IEEEtitleabstractindextext{%
\begin{abstract}
Physical adversarial camouflage poses a severe security threat to autonomous driving systems by mapping adversarial textures onto 3D objects. 
Nevertheless, current methods remain brittle in complex dynamic scenarios, failing to generalize across diverse geometric (e.g., viewing configurations)  and radiometric (e.g., dynamic illumination, atmospheric scattering) variations. 
We attribute this deficiency to two fundamental limitations in simulation and optimization. 
First, the reliance on coarse, oversimplified simulations (e.g., via CARLA) induces a significant domain gap, confining optimization to a biased feature space. 
Second, standard strategies targeting average performance result in a rugged loss landscape, leaving the camouflage vulnerable to configuration shifts.
To bridge these gaps, we propose the Relightable Physical 3D Gaussian Splatting (3DGS) based Attack framework (R-PGA). 
Technically, to address the simulation fidelity issue, we leverage 3DGS to ensure photo-realistic reconstruction and augment it with physically disentangled attributes to decouple intrinsic material from lighting. 
Furthermore, we design a hybrid rendering pipeline that leverages precise Relightable 3DGS for foreground rendering, while employing a pre-trained image translation model to synthesize plausible relighted backgrounds that align with the relighted foreground.
To address the optimization robustness issue, we propose the Hard Physical Configuration Mining (HPCM) module, designed to actively mine worst-case physical configurations and suppress their corresponding loss peaks. This strategy not only diminishes the overall loss magnitude but also effectively flattens the rugged loss landscape, ensuring consistent adversarial effectiveness and robustness across varying physical configurations.
Extensive experiments confirm R-PGA's state-of-the-art performance and superior robustness in both digital and physical domains, where it outperforms the best-competing baselines by further reducing the average AP@0.5 by 6.56\% and 6.12\%, respectively.
Our code is available at: https://github.com/TRLou/R-PGA.

\end{abstract}

}

\maketitle

\IEEEdisplaynontitleabstractindextext

%
\IEEEpeerreviewmaketitle

\section{Introduction}
\label{sec:intro}
Deep Neural Networks (DNNs) have achieved substantial breakthroughs in diverse domains, ranging from computer vision~\cite{he2016resnet} to natural language processing~\cite{vaswani2017attention, fan2023advances}.
However, the advent of adversarial attacks reveals the inherent vulnerabilities of these models. 
While digital attacks~\cite{goodfellow2014FGSM, gu2023survey, carlini2017cw,jia2020adv,he2023generating,lou2024hitadv,jia2025adversarial, muxue2023adversarial,jia2024semantic} across various tasks have sparked widespread security concerns, physical attacks implemented in real-world environments present more severe risks. 
Such threats severely impede the deployment of DNNs in safety-critical fields, including autonomous driving~\cite{wang2023does, cao2023you, deng2020analysis}, security surveillance~\cite{nguyen2023physical, wang2019advpattern, liang2021generate, liang2020efficient, wei2018transferable, liang2022parallel, liang2022large, kong2024patch, liang2024object, kong2024environmental}, and remote sensing~\cite{wang2024fooling, lian2022benchmarking, liu2023x}.
In this paper, we focus on physical attacks in autonomous driving, primarily targeting vehicle detection.

Physical attacks typically involve digitally simulating physical deployment effects and iteratively optimizing perturbations. 
Common implementations include patch application~\cite{eykholt2018rp2, brown2017adversarial, hu2021naturalistic} and camouflage deployment~\cite{zhang2018camou, wu2020genetic, wang2024TAS, zhang2023boosting, suryanto2022DTA, suryanto2023ACTIVE, wang2021DAS, wang2022FCA, zhu2024multiview, liang2025grac, liang2025gcac, zhou2025raucae2e, wang2025highly, zhang2025phycamo}, with the latter becoming a more prevalent research direction due to its higher robustness across different environmental settings.
Unlike adversarial patches, which typically act as localized 2D overlays on the target image, adversarial camouflage necessitates mapping adversarial textures onto 3D surfaces, involving complex geometry-aware computations.
To implement this mapping, early works~\cite{zhang2018camou, wu2020genetic} relied on black-box approximations, whereas subsequent studies~\cite{wang2024TAS, zhang2023boosting, suryanto2022DTA, suryanto2023ACTIVE, wang2021DAS, wang2022FCA, zhu2024multiview, liang2025grac, liang2025gcac, zhou2025raucae2e, wang2025highly, zhang2025phycamo} utilize differentiable neural renderers to enable precise white-box optimization. 
By demonstrating adversarial effectiveness, transferability, and imperceptibility, these works pose a severe security threat to autonomous driving systems.

Despite the progress made by previous methods, the robustness and adversarial effectiveness of the generated camouflage in physical environments remain limited due to limitations in simulation fidelity and optimization objectives: 
(1) regarding scene modeling, existing approaches rely heavily on synthetic environments constructed by simulators (e.g., CARLA), which inevitably deviate from the real physical world; 
(2) physical attributes are often oversimplified, as current methods typically ignore environmental illumination or employ idealized lighting models, while also neglecting the material properties of the camouflage; 
(3) existing optimization strategies primarily focus on average attack performance, yielding a rugged loss landscape with local high-loss peaks in the physical parameter space, thereby making the camouflage extremely sensitive to environmental changes and lacking robustness.

To address these limitations, we propose the \textbf{R}elightable \textbf{P}hysical 3D \textbf{G}aussian Splatting (3DGS) based \textbf{A}ttack framework (R-PGA), which is built upon two core components: a High-Fidelity Relightable Scene Simulator and a Hard Physical Configuration Mining (HPCM) module.
Regarding the simulator, we technically implement it via two key designs. \textbf{First}, we introduce 3DGS as a differentiable high-fidelity renderer in this framework, leveraging its exceptional capability in scene reconstruction and fast, photo-realistic differentiable rendering to support the iterative attack optimization.
Further, we augment the vanilla 3DGS by incorporating intrinsic physical attributes (e.g., albedo, roughness, normal and metallic) and integrating a physically-based rendering (PBR) pipeline, establishing a physically disentangled representation that enables the independent manipulation of surface texture and environmental illumination. 
Crucially, this disentanglement also resolves the cross-view texture inconsistency issue identified in our previous work~\cite{lou2025pga}. 
This inconsistency stems from vanilla 3DGS's reliance on Spherical Harmonics (SH), which entangles lighting with texture to produce view-dependent appearance. 
During the attack optimization, this allows the camouflage to manifest distinct textures for specific viewpoints, thereby preventing convergence onto a single, unified physical camouflage.
\textbf{Second}, global material decomposition proves ill-posed and inefficient, often causing floaters in sparse regions and unnecessary overhead for backgrounds irrelevant to the attack.
We therefore design a hybrid pipeline where the foreground utilizes PBR-based 3DGS for precise physical control, while the background is synthesized via a flow matching-based translation model. 
By inferring environmental contexts from foreground relighting differences, the original background and the target environment map, this approach achieves seamless full-scene relighting without complex background decomposition.
\textbf{Finally}, regarding the optimization, standard optimization yields a rugged loss landscape susceptible to failure peaks across varying physical configurations, involving both shooting parameters (pitch, azimuth, distance) and environmental lighting.
We therefore propose Hard Physical Configuration Mining (HPCM) to actively mine the worst-case physical configurations from a global scope. 
By systematically suppressing these peaks, HPCM effectively flattens the optimization landscape, guaranteeing consistent robustness.
Extensive experiments demonstrate that our attack framework outperforms state-of-the-art methods in both the digital and physical domains. 

Our main contributions are in five aspects:
\begin{itemize}
\item We propose the first physical adversarial attack framework based on 3DGS, which utilizes high-fidelity reconstruction to reduce the digital-physical domain gap and fast differentiable rendering to support efficient iterative optimization.
\item We introduce a physically disentangled representation and a PBR pipeline to decouple surface color from lighting, enabling relighting during optimization and resolving the cross-view texture inconsistency issue.
\item We design a hybrid rendering pipeline combining foreground PBR-3DGS with flow matching-based background translation, which avoids ill-posed global decomposition and achieves seamless full-scene relighting.
\item We propose Hard Physical Configuration Mining (HPCM) to actively mine and suppress worst-case physical configurations, which flattens the rugged loss landscape and guarantees consistent robustness against environmental variations.
\item Extensive experiments demonstrate that R-PGA significantly outperforms state-of-the-art methods in both digital and physical domains, exhibiting superior adversarial effectiveness and robustness against physical configuration variations.

\end{itemize}

This paper is a journal extension of our conference paper~\cite{lou2025pga} (called PGA). This article represents a substantial extension of our preliminary conference version. The main improvements are summarized in the following four aspects:
1) \textbf{Methodology}: Compared to the conference version, this work introduces a Relightable 3DGS framework and a hybrid rendering pipeline, enabling high-fidelity relighting during the optimization process. This advancement significantly enhances the robustness of the generated camouflage against dynamic illumination. Notably, by avoiding SH for surface color representation, we explicitly decouple the lighting information that was previously baked into the texture. This formulation fundamentally resolves the cross-view texture inconsistency issue identified in our conference version by eliminating its root cause. Furthermore, by proposing the HPCM strategy, R-PGA effectively flattens the adversarial loss landscape within the multi-dimensional physical parameter space.
2) \textbf{Experiments}: We have conducted a comprehensive overhaul of all experiments in both digital and physical domains, comparing R-PGA against a broader range of recent SOTA methods. Specifically, we conduct detailed comparative evaluations across four distinct physical configuration dimensions. We also include attack evaluations against two advanced vision foundation models to demonstrate transferability. Additionally, extensive visualization and ablation analyses have been included to facilitate a deeper understanding of the method's efficacy.
3) \textbf{Theory}: We provide a theoretical analysis of the HPCM optimization objective. We prove that HPCM efficiently optimizes the worst-case bound without requiring explicit inner loops. In contrast to standard min-max optimization strategies, our approach avoids intractable computational costs while maintaining theoretical rigor.
4) \textbf{Presentation}: We have completely rewritten the Abstract, Introduction, Method, Experiment, and Conclusion sections to better clarify our motivation and approach. Additionally, we have updated all figures and tables to improve clarity and presentation quality.

\section{Related Work}
\label{related_work}
\subsection{Physical Adversarial Attack}
The landscape of autonomous driving security has been increasingly scrutinized due to the emergence of numerous physical attack techniques targeting critical perception tasks, including traffic sign~\cite{duan2020naturalstyles, feng2021metaattack, eykholt2018rp2, song2018physical_objectdetection}, pedestrian~\cite{sun2023differential, hu2022adversarial, hu2023physically, huang2020universal, xu2020adversarial, thys2019fooling}, and vehicle detection~\cite{zhang2018camou, wu2020genetic, wang2024TAS, zhang2023boosting, suryanto2022DTA, suryanto2023ACTIVE, wang2021DAS, wang2022FCA, zhu2024multiview, liang2025grac, liang2025gcac, zhou2025raucae2e, wang2025highly, zhang2025phycamo}.
Adversaries typically optimize adversarial patches, clothes, and camouflage, among which camouflage holds greater practical value as it remains effective across diverse viewing configurations.
The optimization of camouflage hinges on rendering textures onto target surfaces, such as vehicles. As early explorations, several studies adopted black-box strategies to tackle the non-differentiable rendering process. 
Specifically, Zhang et al.~\cite{zhang2018camou} trained a neural network to approximate the rendering function, while Wu et al.~\cite{wu2020genetic} leveraged a genetic algorithm to directly search for optimal adversarial camouflage.
To exploit white-box settings for enhanced adversarial capabilities, recent studies~\cite{wang2021DAS, wang2022FCA, suryanto2022DTA, suryanto2023ACTIVE, wang2024TAS} have utilized differentiable rendering methods~\cite{Kato2018NMR, suryanto2022DTA}. 
Wang et al.~\cite{wang2021DAS} proposed suppressing both model and human attention to ensure visual naturalness, and later extended this to model-shared attention for better transferability~\cite{wang2024TAS}. 
Addressing partial occlusion and long-distance detection, Wang et al.~\cite{wang2022FCA} optimized full-coverage vehicle camouflage. 
Meanwhile, Suryanto et al.~\cite{suryanto2022DTA} integrated a photo-realistic renderer to boost robustness, further improving universality via tri-planar mapping~\cite{suryanto2023ACTIVE}.
Focusing on the optimization process, Liang et al.~\cite{liang2025gcac} introduced gradient calibration and decorrelation strategies to resolve inconsistent sampling densities and conflicting multi-view updates.

Despite advancements in cross-view and cross-distance robustness, previous approaches still suffer from sensitivity to illumination and weather.
Zhou et al.~\cite{zhou2024rauca} addressed this by integrating an environment feature extractor to simulate diverse conditions, and later introduced end-to-end UV map optimization to minimize sampling errors~\cite{zhou2025raucae2e}.
Similarly, Liang et al.~\cite{liang2025grac} proposed the GRAC framework, which models light interactions and employs gradient reweighting to enhance robustness.
Liu et al.~\cite{liu2025naturalistic} proposed a dual-constraint framework that employs global illumination-based rendering to model physical optical interactions and a GAN-based style learner to ensure visual plausibility.

\subsection{Advanced 3D Representations for Physical Attacks}
Most of the above works rely on low-fidelity simulations of target objects and environments. 
The inherent coarse and simplified modeling inevitably deviates from real-world scenarios, leading the optimization into a biased feature space and ultimately yielding sub-optimal solutions.
Recently, advanced 3D representations, such as NeRF~\cite{mildenhall2021nerf} and 3D Gaussian Splatting ~\cite{kerbl20233dgs}, have facilitated the modeling of objects and scenes, offering differentiable rendering pipelines applicable to physical attack frameworks.
Li et al.~\cite{li2023adv3d} represented target vehicles via NeRFs to optimize adversarial patches, achieving improved physical realism. Huang et al.~\cite{huang2024tt3d} proposed a transferable targeted attack utilizing grid-based NeRF for mesh reconstruction, simultaneously optimizing texture and geometry. 
While these approaches circumvent the reliance on simulation software and enable direct modeling of real-world scenes, these methods remain constrained by NeRF's inherent limitations, such as slow rendering, limited fidelity, and high memory consumption.
Alternatively, 3D Gaussian Splatting (3DGS) provides a promising solution by utilizing differentiable splatting operations, enabling rapid and high-fidelity rendering suitable for iterative adversarial optimization
Leveraging these advantages, Lou et al.~\cite{lou2025pga} introduced the first 3DGS-based attack framework, which effectively addresses mutual and self-occlusion among Gaussians and enhances robustness via pixel-perturbation-based min-max optimization.
Despite achieving high-fidelity modeling and rendering results, these works fail to support the simulation of lighting and weather variations, leading to limited camouflage robustness.

The capability to support relighting and material decomposition in 3D Gaussian Splatting (3DGS) has emerged as a common prerequisite for various tasks. 
Contemporary approaches~\cite{shi2025gir, moenne2024raytracing, liang2024gsir, gao2024relightable, ye2025geosplatting, jiang2024gaussianshader, kaleta2025lumigauss} generally extend the attributes of Gaussian ellipsoids by additionally learning physical properties such as normals, roughness, metallic, and albedo. 
By integrating these properties with diverse lighting models to compute direct and indirect illumination, these methods achieve Physically Based Rendering (PBR). 
In this work, we leverage relightable 3DGS as the backbone of our rendering pipeline, introducing specific adaptations to tailor it for physical camouflage generation scenarios. Coupled with enhanced optimization strategies, we present a framework capable of generating camouflage that is robust to both geometric and radiometric variations.

\section{Preliminaries}
\label{preliminaries}
This section introduces 3D Gaussian Splatting (3DGS) and the physical adversarial attack formulation, followed by an analysis of prior limitations in simulation fidelity and optimization objectives.

\subsection{3D Gaussian Splatting}
3DGS reconstructs the scene by representing it with a large set of Gaussians $\mathcal{G} = \{\bm{g}_1, \bm{g}_2, ..., \bm{g}_N\}$, where $N$ denotes the number of Gaussians.
Each Gaussian $\bm{g}$ is characterized by its mean $\bm{\mu_g}$ and anisotropic covariance $\bm{\Sigma_g}$, and can be mathematically represented as:
\begin{equation}
\vspace{-1mm}
\bm{g}(\bm{x}) = exp(-\frac{1}{2}(\bm{x}-\bm{\mu_g})^T\bm{\Sigma}_g^{-1}(\bm{x}-\bm{\mu_g})),
\label{eq:}
\end{equation}
where the mean $\bm{\mu_g}$ determines its central position, and the covariance $\bm{\Sigma}_g$ is defined by a scaling vector $\bm{s}_g \in \mathbb{R}^3$ and a quaternion $\bm{q}_g \in \mathbb{R}^4$ that encodes the rotation of $\bm{g}$.
Besides, 3DGS uses an $\bm{\alpha_g} \in [0, 1]$ to represent the opacity of $\bm{g}$ and describes the view-dependent surface color $\bm{c}_g$ through spherical harmonics coefficients $\bm{k}_g$. 
To reconstruct a new scene, 3DGS requires only a few images $\mathcal{I}$ from different viewpoints as training inputs. 
Starting from a point cloud initialized by SfM~\cite{noah2006sfm}, it optimizes and adjusts the parameters $\{\bm{\mu_g}, \bm{s_g}, \bm{q_g}, \bm{\alpha_g}, \bm{k_g}\}$ of each $\bm{g}$ to make the rendering closely resemble the real images.
After training, an image $\bm{I_{\theta_c}}$ can be differentially rendered through a rasterizer $\mathcal{R}$ by splatting each 3D Gaussian $\bm{g}$ onto the image plane as a 2D Gaussian, with pixel values efficiently computed through alpha blending given a viewpoint $\bm{\theta_c}$ and a set $\mathcal{G}$, formulated as $\bm{I_{\theta_c}} = \mathcal{R}(\mathcal{G}, \bm{\theta_c})$.

\subsection{Formulation of Physical Attack}

The primary goal of physical adversarial attacks is to generate robust adversarial camouflage $\bm{\mathcal{T}}$ that remains effective across the distribution of real-world viewing and environmental conditions, denoted as $\mathcal{D}_{\text{real}}$.
Since directly optimizing over the non-differentiable and complex $\mathcal{D}_{\text{real}}$ is intractable, physical attack methods employ rendering pipelines to synthesize the detection input images $\mathcal{I}_{\text{det}}$ by compositing the rendered foreground $\mathcal{R}(\bm{\mathcal{T}}, \bm{c})$ with the background $\mathcal{B}$.
Here, $\bm{c}=(\bm{\phi},\bm{\theta},\bm{d},\bm{E})\in\mathcal{C}_{\text{sim}}$ represents the physical configuration, sampled from the simulator configuration space $\mathcal{C}_{\text{sim}}$, including camera pitch, azimuth, shooting distance, and environment map which is used in image-based lighting.
These synthesized data constitute a simulated distribution $\mathcal{D}_{\text{sim}}$.
Consequently, the discrepancy between $\mathcal{D}_{\text{sim}}$ and $\mathcal{D}_{\text{real}}$ formally characterizes the domain gap between the digital and physical worlds.

To ensure the camouflage is robust against these variations, the generation of $\bm{\mathcal{T}}$ is typically formulated as an Expectation over Transformations (EoT) optimization problem. 
The objective is to minimize the expected adversarial loss over the distribution of physical configurations:

\begin{equation}
\min_{\bm{\mathcal{T}}} \mathbb{E}_{\bm{c} \sim \mathcal{C}} \left[ \mathcal{L}_{\text{adv}}(\mathcal{F}(\mathcal{R}(\bm{\mathcal{T}},\bm{c})+\mathcal{B}), y) \right],
\label{eq:standard_eot}
\end{equation}
where $\mathcal{F}$ denotes the victim detector and $y$ represents the ground truth label.
This formulation aims to find the optimal $\bm{\mathcal{T}}$ that minimize the average detection performance of $\mathcal{F}$ across the simulated space.

\subsection{Problem Analysis}
\label{sec:3.3}
Although the formulation in Eq.~\ref{eq:standard_eot} is widely adopted, we observe that adversarial camouflage generated by prior methods often exhibits insufficient robustness and adversarial effectiveness in practice. 
We attribute these performance limitations to the discrepancies between the simplified simulation and the complex physical world, as well as the pitfalls of the optimization strategy itself. 
Formally, we characterize the underlying causes of these deficiencies as two fundamental gaps:
\begin{itemize}

\item \textbf{The Domain and Configuration Gap:} There exists a distributional shift between the rendered images $\mathcal{I}_{\text{det}} \sim \mathcal{D}_{\text{sim}}$ (where $\mathcal{I}_{\text{det}} = \mathcal{F}(\mathcal{R}(\bm{\mathcal{T}}, \bm{c})+\mathcal{B})$) and real-world captures $\mathcal{I}_{\text{real}} \sim \mathcal{D}_{\text{real}}$.
Furthermore, the simulated configuration space $\mathcal{C}_{\text{sim}}$ deviates from the real-world space $\mathcal{C}_{\text{real}}$; specifically, $\mathcal{C}_{\text{sim}}$ typically lacks the illumination dimension or models it in an oversimplified manner.

\item \textbf{The Optimization Objective Gap:} Prior methods minimize the expected loss $\mathbb{E}_{\bm{c} \sim \mathcal{C}}[L(\cdot)]$, which inherently ignores the loss variance. This formulation permits failure peaks in the physical parameter space, resulting in a rugged loss landscape that undermines robustness against configuration shifts.

\end{itemize}

\section{Methodology}

\begin{figure*}[t]
  \centering
   \includegraphics[width=1\linewidth]{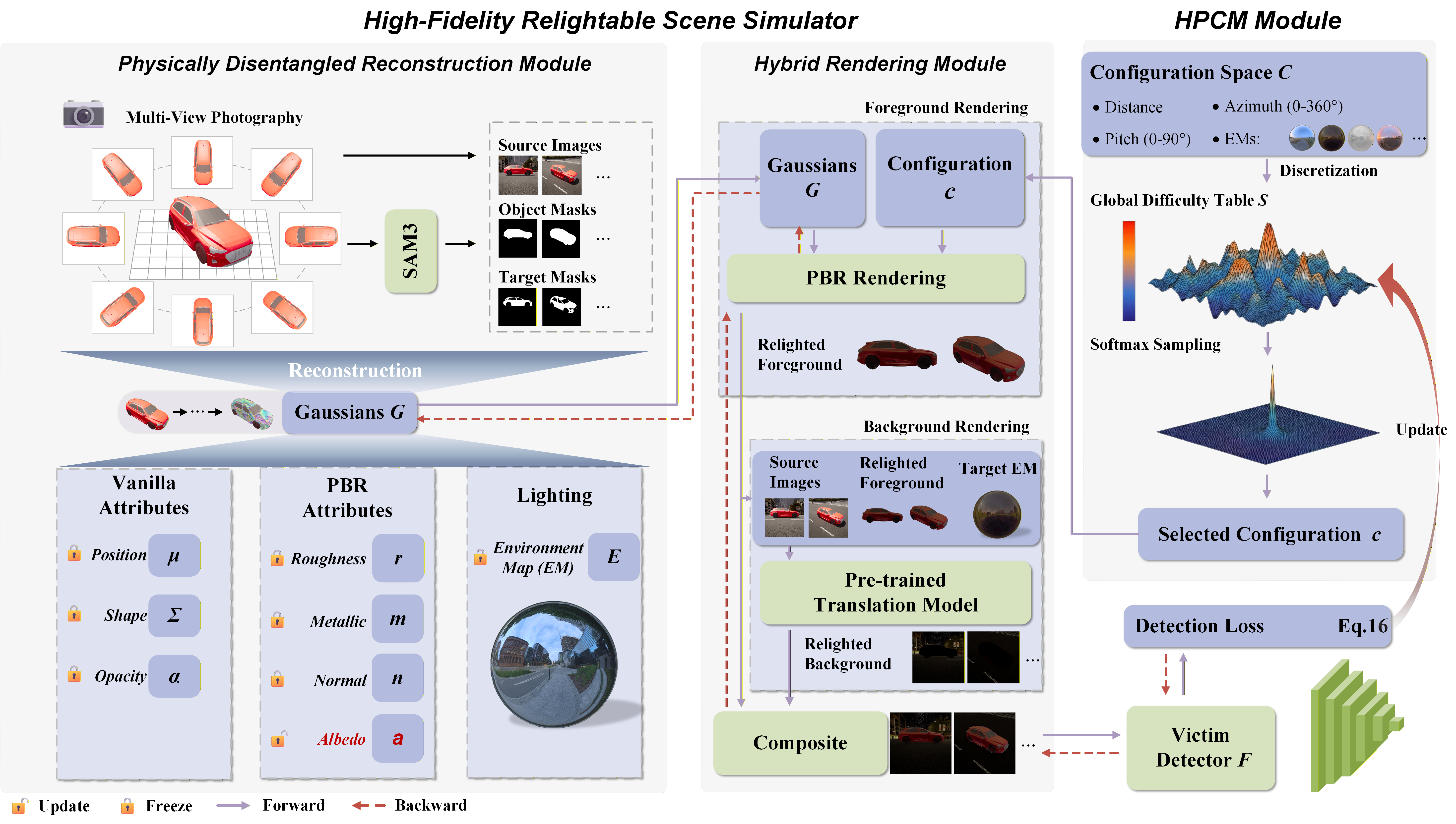}

   \caption{Demonstration of the framework of R-PGA. It consists of the High-Fidelity Relightable Scene Simulator and the HPCM Module. The simulator features a reconstruction and rendering pipeline composed of the Physically Disentangled Reconstruction Module and the Hybrid Rendering Module. The HPCM module constructs a discretized Configuration Space to build a Global Difficulty Table, which is utilized to select physical configurations during each iteration.
}
   \label{fig:framework}
   \vspace{-2mm}
\end{figure*}

\subsection{Overview}
\label{sec:4.1}
To address the two critical issues identified in Sec.~\ref{sec:3.3} that degrade adversarial effectiveness and robustness, we propose R-PGA, a novel physical attack framework based on relightable 3DGS.
R-PGA comprises two core components: the High-Fidelity Relightable Scene Simulator, which provides physically decoupled, high-fidelity scene reconstruction and fast differentiable rendering to bridge the domain and configuration gap; 
and the Hard Physical Configuration Mining module, which guides the generated adversarial camouflage towards a flatter region within the physical parameter space to address the optimization objective gap.
Leveraging these two core components, we establish the R-PGA framework to iteratively refine the attributes of the Gaussians $\mathcal{G}$, yielding robust adversarial Gaussians $\mathcal{G}'$.
Subsequently, the adversarial camouflage $\bm{\mathcal{T}}$ is extracted from $\mathcal{G}'$ using the method proposed in \cite{guedon2024sugar} to mislead the detector $\mathcal{F}$.
We elaborate on the two components in Sec.~\ref{sec:4.2} and Sec.~\ref{sec:4.3}, respectively, and detail the loss function design and implementation specifics in Sec.~\ref{sec:4.4}.
In Sec.~\ref{sec:4.5}, we provide a theoretical analysis of the strategy design of HPCM.
The overall framework is illustrated in Fig.~\ref{fig:framework}.

\subsection{High-Fidelity Relightable Scene Simulator}
\label{sec:4.2}
\subsubsection{Physically Disentangled Reconstruction Module}
\label{sec:reconstruction}

\begin{figure}[t]
  \centering
   \includegraphics[width=0.75\linewidth]{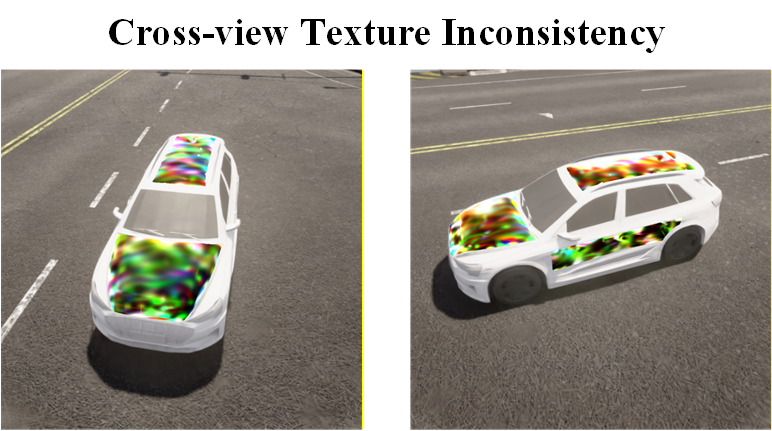}

   \caption{Illustration of cross-view texture inconsistency observed when generating adversarial camouflage using vanilla 3DGS. This issue stems from the fact that lighting information, which is baked into the Spherical Harmonics (SH) coefficients and should remain invariant, is erroneously modified during the iterative optimization process.
}
   \label{fig:inconsistency}
\end{figure}

To enable robust physical attacks, the adversarial perturbations must reflect the intrinsic surface properties rather than transient lighting effects. 
Standard 3DGS bakes lighting into view-dependent Spherical Harmonics, causing texture inconsistencies across different viewing angles that destabilize the iterative attack optimization, as shown in Fig.~\ref{fig:inconsistency}. 
To fundamentally resolve this, we augment the Gaussian primitives with explicit physically-based rendering (PBR) attributes: albedo $\mathbf{a} \in [0,1]^3$, metallic $\bm{m} \in [0,1]$, normal $\bm{n}$ and roughness $\bm{r} \in [0,1]$.
Crucially, optimizing the albedo $\mathbf{a}$ allows us to generate adversarial patterns that represent the surface's inherent color, free from the interference of baked illumination, thereby ensuring consistency across diverse physical configurations.

We replace the standard color rendering with a physically-based shading model. Specifically, the outgoing radiance $L_o$ from a 3D Gaussian at position $\mathbf{x}$ along the viewing direction $\boldsymbol{\omega}_o$ is computed via the rendering equation:
\begin{equation}
    L_o(\boldsymbol{\omega}_o, \mathbf{x}) = \int_{\Omega} f_r(\boldsymbol{\omega}_i, \boldsymbol{\omega}_o, \mathbf{x}) L_i(\boldsymbol{\omega}_i, \mathbf{x}) (\boldsymbol{\omega}_i \cdot \mathbf{n}) d\boldsymbol{\omega}_i,
    \label{eq:rendering_eq}
\end{equation}
where $\boldsymbol{\omega}_i$ denotes the incident light direction, and $L_i$ corresponds to the incident radiance. $f_r$ represents the BRDF properties of the surface. The integration domain is the upper hemisphere $\Omega$ defined by the point $\mathbf{x}$ and its surface normal $\mathbf{n}$.
For the BRDF term $f_r$, we employ the widely adopted Disney Micro-facet model~\cite{burley2012disney}, which utilizes the optimized $\bm{m}$ and $\bm{r}$ to accurately model surface reflections.
To solve Eq.~\eqref{eq:rendering_eq} efficiently while maintaining high fidelity, we adopt the hybrid illumination and geometry estimation strategy established in GIR~\cite{shi2025gir}. 
Specifically, the incident light $L_i$ is decomposed into direct and indirect components:
(1) Direct Lighting: Represented by a high-resolution environment map $\mathbf{E}$ using Image-Based Lighting (IBL).
(2) Indirect Lighting: Modeled via Spherical Harmonics to capture multi-bounce effects, modulated by a visibility term to handle occlusions.
Consistent with~\cite{shi2025gir}, the surface normal $\mathbf{n}$ is derived from the shortest axis of the Gaussian's covariance matrix, ensuring geometric plausibility without external supervision.

Prior to generating adversarial camouflage, we first acquire a set of multi-view ground truth images $\mathcal{I}_{\text{gt}} = \{\bm{I}_1, \bm{I}_2, \dots\}$ of the target object.
We then reconstruct a set of 3D Gaussians $\mathcal{G} = \{\bm{g}_1, \bm{g}_2, \dots\}$, and optimize their attributes by minimizing the pixel-wise difference between the rendered and ground truth images. 

\subsubsection{Hybrid Rendering Module}
In our previous conference work, PGA~\cite{lou2025pga}, we employed 3DGS to reconstruct the entire scene, utilizing a target mask to restrict the update of Gaussian attributes to the target object's surface. 
However, in R-PGA, we observe that attempting to reconstruct and perform material decomposition on the entire scene leads to a collapse in reconstruction quality.
Specifically, given that the scene resides in an open environment, peripheral regions suffer from sparse viewpoint supervision. 
This results in significant geometric errors, ill-posed material and lighting decomposition, artifacts, and floaters. 
Crucially, these errors mutually exacerbate one another, leading to a total failure of the reconstruction.

Given our focus on physical adversarial attacks, high-quality reconstruction and decomposition of the scene periphery yields negligible benefits, as we do not need to iteratively manipulate these regions, while introducing substantial computational complexity and overhead.
Therefore, we propose a hybrid rendering framework: for the critical target object, we employ a high-precision, multi-view supervised Relightable 3DGS reconstruction; for the background, we utilize 2D image translation to adapt to lighting changes in the foreground and generate backgrounds that align with the real data distribution.

Specifically, during the relightable 3DGS reconstruction, we first utilize the Segment Anything Model (SAM) \cite{kirillov2023sam} to extract the target object from the collected multi-view images with object masks $\mathcal{M}_{\text{obj}}$: 

\begin{equation}
\mathcal{I}_{\text{gt}}^{\text{obj}} = \text{SAM}(\mathcal{I}_{\text{gt}}, \mathcal{P}_{\text{obj}}) \odot \mathcal{I}_{\text{gt}}=\mathcal{M}_{\text{obj}} \odot \mathcal{I}_{\text{gt}},
\label{eq:}
\end{equation}
where $\mathcal{P}_{\text{obj}}$ are prompts of the target object.
This enables us to exclusively reconstruct the 3D Gaussians of the target object with $\mathcal{I}_{\text{gt}}^{\text{obj}}$, denoted as $\mathcal{G}_{\text{obj}} = \{g_1, g_2, \dots, g_N\}$, where $N$ represents the total number of Gaussians.
Subsequently, we formulate the foreground rendering process as follows:

\begin{equation}
\mathcal{I}_{\text{fg}} = \mathcal{R}(\mathcal{G}_{\text{obj}}, c) \odot \mathcal{M}_{\text{tar}} + \mathcal{I}_{\text{gt}}^{\text{obj}} \odot (\mathcal{M}_{\text{obj}}-\mathcal{M}_{\text{tar}}),
\label{eq:}
\end{equation}
where $\mathcal{R}$ denotes the rasterizer provided by 3DGS, and $\mathcal{M}_{\text{tar}}=\text{SAM}(\mathcal{I}_{\text{gt}}, \mathcal{P}_{\text{tar}})$ represents the target mask corresponding to the camouflage region, which is defined by pre-applying stickers with specific patterns (e.g., red stickers) on the object.

As for background rendering, we employ a pre-trained image translation model LBM~\cite{chadebec2025lbm} to synthesize high-fidelity backgrounds that are physically consistent with the relighted foreground.
To support this, we construct a multi-illumination dataset using the CARLA simulator and train the LBM model to capture the correlation between environmental lighting and scene appearance (detailed training configurations and dataset construction are provided in the Appendix).

During the rendering process, the LBM acts as a conditional generator.
It takes the target environment map $\bm{E}$, the ground truth image $\mathcal{I}_{\text{gt}}$ and the photometric cues from the rendered foreground $\mathcal{I}_{\text{fg}}$ (produced by our Relightable 3DGS) as inputs. The model then infers the corresponding background $\mathcal{I}_{\text{bg}}$, ensuring that the background illumination naturally aligns with the foreground's lighting conditions. Formally, the background generation is defined as:

\begin{equation}
\mathcal{I}_{\text{bg}} = \mathcal{H}(\mathcal{I}_{\text{fg}}, \mathcal{I}_{\text{gt}}, \bm{E}),
\label{eq:}
\end{equation}
where $\mathcal{H}$ denotes the pre-trained LBM inference function. Finally, the complete scene is composited using the mask $\mathcal{M}$:

\begin{equation}
\mathcal{I}_{\text{det}} = \mathcal{M}_{\text{obj}} \odot \mathcal{I}_{\text{fg}} + (1 - \mathcal{M}_{\text{obj}}) \odot \mathcal{I}_{\text{bg}}.
\label{eq:}
\end{equation}
This generative approach effectively bypasses the ill-posed problem of decomposing background materials in open scenes, yielding a seamless and realistic composite for physical adversarial optimization.

\begin{figure*}[t]
  \centering
   \includegraphics[width=1\linewidth]{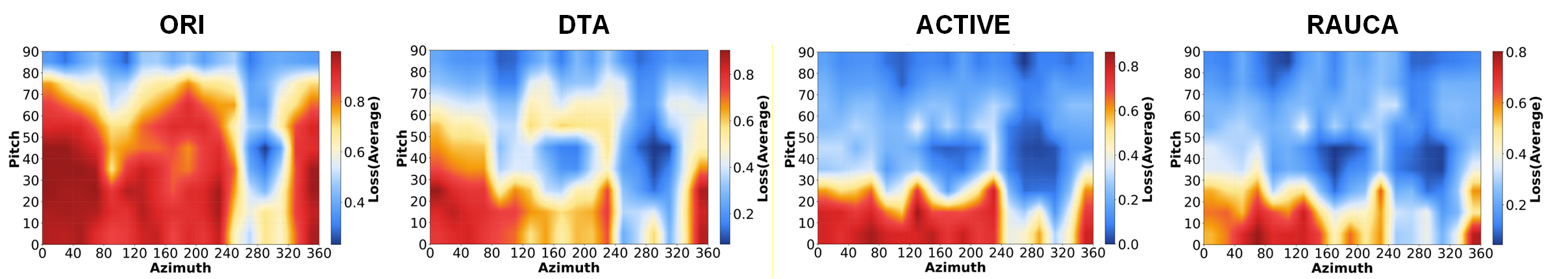}

   \caption{Detection heatmaps (Red: Detected, Blue: Evaded) across Azimuth and Pitch angles, averaged over other configuration dimensions and multiple detectors, and three EoT-based SOTA methods exhibit similar failure regions. 
}
   \label{fig:heatmap}
   \vspace{-4mm}
\end{figure*}

\subsection{Hard Physical Configuration Mining Module}
\label{sec:4.3}
\textbf{Motivation.}
To investigate the distribution of attack robustness across the physical parameter space, we visualize the adversarial loss landscape in Fig.~\ref{fig:heatmap}.
We observe a phenomenon of spatial consistency: regardless of the specific attack pattern or initialization, regions of high adversarial loss (indicating attack failure) consistently cluster around specific physical configurations.
We term this property inherent configuration hardness, which stems from the intrinsic geometric structures and non-camouflaged regions of the target object, rendering certain configurations naturally resilient to adversarial perturbations.
Under these challenging configurations, the camouflage requires extensive iterative optimization to converge to a better solution. However, the standard EoT framework treats all configurations uniformly.
Consequently, the optimization of these hard configurations is disrupted by gradients from easier views at the early stages, leading to premature convergence to local optima.
Ultimately, this imbalance creates a rugged loss landscape in the physical parameter space, characterized by loss peaks at these difficult configurations. 
Therefore, we require an optimization strategy that can adaptively identify and suppress these peaks to achieve more robust camouflage.

\textbf{Strategy Formulation.}
Based on this motivation, we propose the Hard Physical Configuration Mining (HPCM) strategy.
Unlike transient hard mining methods that rely on instantaneous batch losses, HPCM is designed to identify configurations that remain consistently challenging throughout the iterative optimization of the adversarial texture.
We discretize the continuous physical configuration space $\mathcal{C}$ into $q$ distinct bins and maintain a Global Difficulty Table $\mathbf{S} = \{s_1, s_2, \dots, s_q\}$ to track the historical robustness of each configuration.
To encourage the optimizer to explore the entire configuration space comprehensively in the early stages, we initialize all entries in $\mathbf{S}$ to a high constant value (e.g., $10$).

During the optimization at step $t$, the difficulty score $s_i$ for a sampled configuration $\bm{c}_i$ is updated.
It is important to note that our goal is to solve for a \textit{universal} adversarial perturbation; thus, the camouflage texture changes dynamically at every iteration. 
Consequently, the instantaneous loss $\mathcal{L}_{\text{curr}}$ reflects only a transient snapshot of the current texture's performance.
To obtain a stable, macroscopic assessment of the configuration's inherent hardness, we employ a momentum-based update mechanism:
\begin{equation}
    s_i^{(t)} = \mu \cdot s_i^{(t-1)} + (1 - \mu) \cdot \mathcal{L}_{\text{curr}},
    \label{eq:momentum_update}
\end{equation}
where $\mu \in [0, 1)$ is a momentum coefficient.
This temporal smoothing accumulates historical gradients, ensuring that high $S_i$ values reflect configurations that are consistently difficult to attack throughout the optimization trajectory.

Leveraging this global difficulty information, we replace uniform sampling with a difficulty-aware sampling mechanism.
In each iteration, the probability $P(c_i)$ of sampling a configuration $c_i$ is proportional to its difficulty score, formulated via a Softmax distribution:
\begin{equation}
    P(c_i) = \frac{\exp(s_i / \tau)}{\sum_{j=1}^{M} \exp(s_j / \tau)},
    \label{eq:sampling_prob}
\end{equation}
where $\tau$ is a temperature hyperparameter.
The utilization of HPCM ensures that the peaks of the loss landscape are adaptively suppressed, progressively achieving uniform robustness.

\subsection{Total Loss and Optimization Objective}
\label{sec:4.4}
In R-PGA, we feed the hybrid rendered images $\mathcal{I}_{\text{det}}$ into the victim white-box detector $\mathcal{F}$ to obtain the detection results:
\begin{equation}
\mathcal{B} = \mathcal{F}(\mathcal{I}_{\text{det}}; \boldsymbol{\theta}_{\mathcal{F}}) = \{\boldsymbol{b}_1, \boldsymbol{b}_2, \dots \}. 
\end{equation}

Subsequently, the detection loss is defined following~\cite{hu2023physically, lou2025pga} as:

\begin{equation}
\begin{aligned}
\mathcal{L}_{\text{det}}(\mathcal{I}_{\text{det}}) &= \sum_{\bm{I}} \text{Conf}_{m^*}^{(\bm{I})}, \\m^*&=\underset{m}{\text{argmax}}
 \text{IoU}(\bm{gt^{(I)}},\bm{b^{(I)}}_m),
\label{eq:}
\end{aligned}
\end{equation}
where $I$ represents each input image in the batch, $\boldsymbol{b}_m$ denotes the $m$-th bounding box in the detection results, and $\text{Conf}$ indicates the confidence score of the corresponding class. $\mathcal{L}_{\text{det}}$ minimizes the confidence of the correct class for the box that has the maximum Intersection over Union (IoU) with the ground truth $\boldsymbol{gt}$.
Simultaneously, we employ the HPCM strategy during optimization to perform hardness-aware sampling. Therefore, the total optimization objective can be formulated as:

\begin{equation}
\mathcal{J} = \mathbb{E}_{\xi \sim \mathcal{D}_{\text{HPCM}}} \left[ \mathcal{L}_{\text{det}}\left( \mathcal{R}(\mathcal{G}(\mathbf{a}), \xi) \right) \right],   
\end{equation}
where $\mathcal{D}_{\text{HPCM}}$ represents the sampling distribution guided by configuration hardness, and $\mathbf{a}$ denotes the view-independent albedo of the 3D Gaussians. Finally, the albedo $\mathbf{a}$ is updated iteratively via gradient descent with a learning rate $\eta$:
\begin{equation}
\mathbf{a}^{t+1} = \mathbf{a}^{t} - \eta \nabla_{\mathbf{a}} \mathcal{J}.
\end{equation}

\begin{table*}[t]
\centering
\caption{Comparison of AP@0.5 for different physical attack methods against various detection models. The reported results represent the average detection performance on the test dataset collected across multiple viewpoints, shooting distances, and weather conditions. Note that the adversarial camouflage is generated using Yolo-V3 and evaluated for black-box transferability (marked with *) on Yolo-X, Faster R-CNN, Mask R-CNN, Deformable-DETR and PVT.}

\rowcolors{3}{white}{gray!10}
\vspace{2mm}
\scalebox{1.}{
\begin{tabular}{@{}c|cc|cc|cc|c@{}}
\toprule
\multirow{2}{*}{Method} & \multicolumn{2}{c|}{One-Stage}    & \multicolumn{2}{c|}{Two-Stage}    & \multicolumn{2}{c|}{Transformer-based} & \multirow{2}{*}{Average} \\ \cmidrule(lr){2-7}
                          & Yolo-V3          & YoloX* & FrRCN* & MkRCN* & D-DETR* & PVT* &                          \\ \midrule
ORI                       & 0.5251          & 0.7395          & 0.5712          & 0.6270          & 0.5877             & 0.7312            & 0.6126                    \\
FCA                       & 0.4982          & 0.6550          & 0.3511          & 0.4035          & 0.4070             & 0.6875            & 0.4758                    \\
ACTIVE                    & 0.1934          & 0.3749          & 0.1460          & 0.1841          & 0.2175             & 0.4160            & 0.2376                    \\
DTA                       & 0.3462          & 0.4069          & 0.2292          & 0.3143          & 0.2924             & 0.5600            & 0.3240                    \\
RAUCA                     & 0.1869          & 0.3643          & 0.1474          & 0.1725          & 0.1059             & 0.5081            & 0.2256                    \\

GCAC                      & 0.1207          & 0.3084          & 0.1216          & 0.1614          & 0.0949             & 0.3556            & 0.1793                    \\
GRAC                      & 0.1515          & 0.3695          & 0.1389          & 0.1888          & 0.1052             & 0.4563            & 0.2160                    \\
RAUCA-E2E                 & 0.1121          & 0.1303          & 0.0256 & 0.1305          & 0.0389             & 0.4041            & 0.1333                    \\
PGA & 0.0318 & 0.3022 & 0.0623 & 0.1530 & 0.0562 & 0.3382 & 0.1572
\\
R-PGA                     & \textbf{0.0227} & \textbf{0.0325} & \textbf{0.0223}          & \textbf{0.0565} & \textbf{0.0206}    & \textbf{0.2521}   & \textbf{0.0677}          \\ \bottomrule
\end{tabular}
}

\label{tab:main}
\end{table*}

\begin{figure*}[h]
  \centering
   \includegraphics[width=0.7\linewidth]{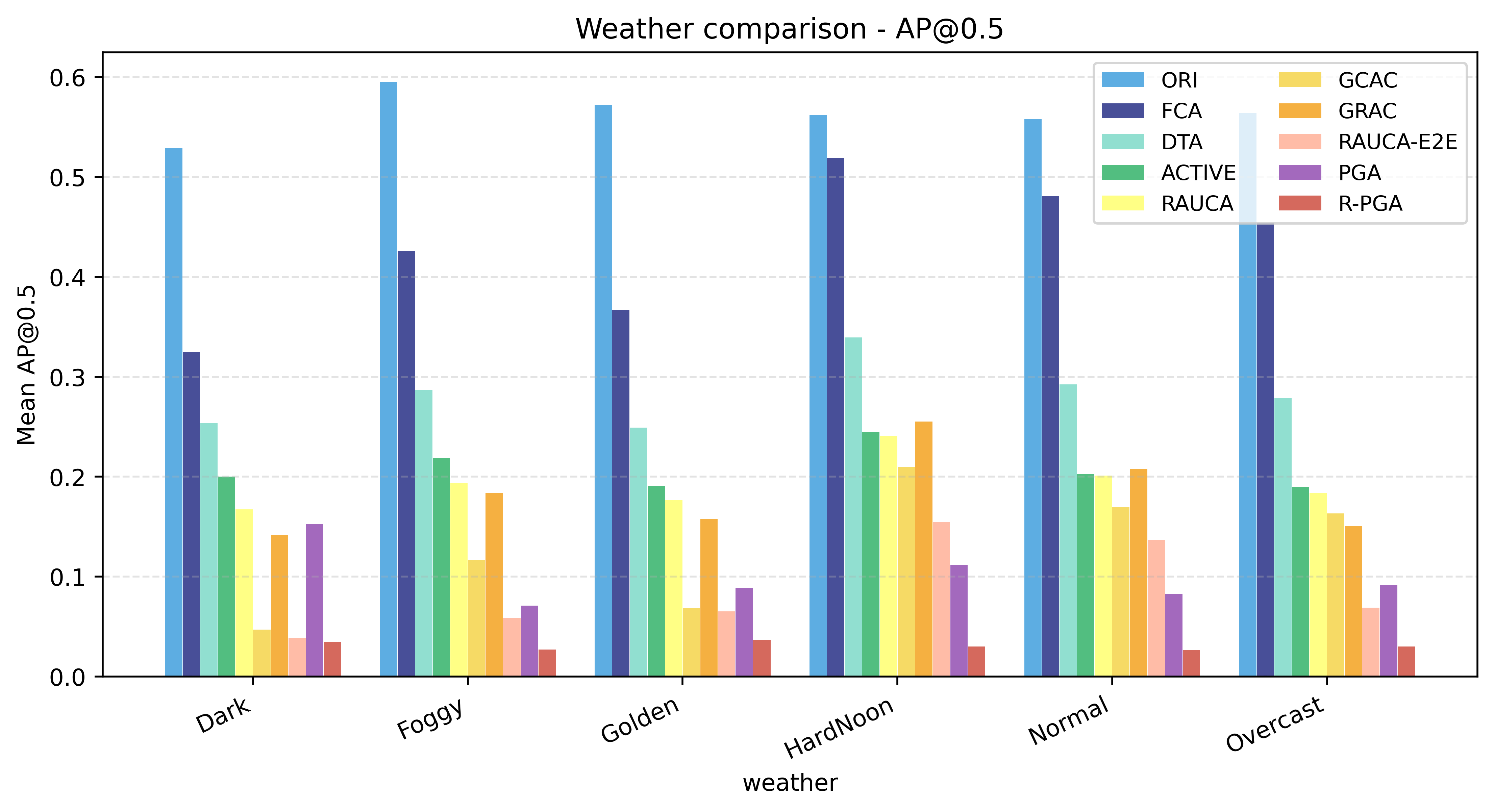}

   \caption{Comparison of detection results for different attack methods across various weather conditions, specifically reporting the average AP@0.5 for different detectors averaged over diverse shooting distances and pitch angles.
}
   \label{fig:weather}
   \vspace{-4mm}
\end{figure*}

\subsection{Theoretical Analysis}
\label{sec:4.5}
To provide a rigorous justification for the proposed Hard Physical Configuration Mining (HPCM), we analyze its underlying optimization objective from the perspective of robust optimization.

\textbf{Min-Max Formulation. }
Ideally, to ensure consistent robustness across all potential physical variations, the attack should aim to minimize the worst-case loss rather than the average performance.
This corresponds to the Min-Max optimization problem:

\begin{equation}
\min_{\mathbf{a}} \max_{c \in \mathcal{C}} \mathcal{L}(\mathbf{a}, c).
\end{equation}
However, directly solving the inner maximization $\max_{c} \mathcal{L}$ is computationally intractable, as it requires an expensive iterative search over the complex rendering pipeline at every training step. 
If limited iterations are used, it becomes difficult to ensure that the true maximum is located, which consequently undermines the efficacy of the bi-level optimization.

\textbf{Efficient Surrogate via Log-Sum-Exp. }To bypass the costly inner maximization while maintaining focus on hard examples, we propose minimizing the Log-Sum-Exp (LSE) function. 
The LSE serves as a smooth upper bound of the maximum function:

\begin{equation}
   \mathcal{J}_{\text{LSE}}(\mathbf{a}) = \tau \log \left( \sum_{i=1} \exp \left( \frac{\mathcal{L}(\mathbf{a}, \bm{c}_i)}{\tau} \right) \right), 
\end{equation}
where $\tau$ is the temperature parameter.

\textbf{Equivalence of HPCM and LSE Optimization.} We formally prove that the sampling strategy employed in HPCM is mathematically equivalent to performing Stochastic Gradient Descent (SGD) on this $\mathcal{J}_{\text{LSE}}$ objective. 
Specifically, the expected gradient with respect to albedo $\mathbf{a}$ under the HPCM sampling distribution $P(\bm{c})$ (Eq.~\ref{eq:sampling_prob}) aligns exactly with the gradient of the LSE function:

\begin{equation}
  \mathbb{E}_{\bm{c} \sim P}[\nabla_{\mathbf{a}} \mathcal{L}(\mathbf{a}, \bm{c})] \equiv \nabla_{\mathbf{a}} \mathcal{J}_{\text{LSE}}(\mathbf{a}).  
\end{equation}
This equivalence implies that HPCM efficiently optimizes the worst-case bound without explicit inner loops. 
The detailed proof is provided in Appendix.

\textbf{Landscape Flattening.} By minimizing the LSE objective, R-PGA implicitly exerts a strong suppression force on the configurations with the highest losses (peaks).
As demonstrated in~\cite{boyd2004convex}, the LSE function strictly upper-bounds the maximum loss. 
Consequently, optimizing this bound reduces the gap between the worst-case and average performance. 
This mechanism effectively flattens the rugged loss landscape, eliminating failure peaks and guaranteeing uniform robustness.

\begin{figure}[t]
  \centering
   \includegraphics[width=1\linewidth]{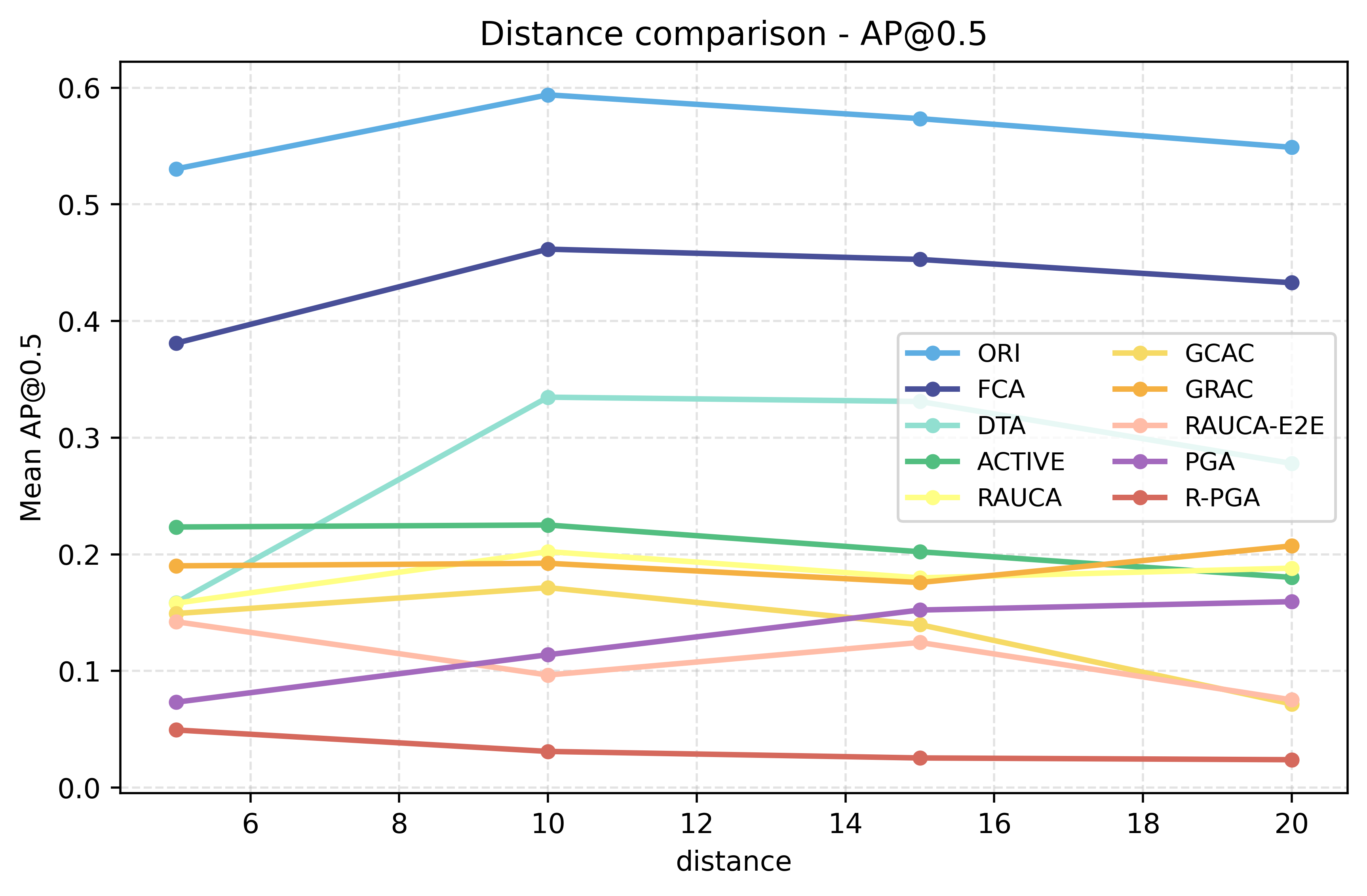}

   \caption{Comparison of detection results for different attack methods across various shooting distances, specifically reporting the average AP@0.5 for different detectors averaged over diverse weather conditions and pitch angles.
}
   \label{fig:distance}
   \vspace{-4mm}
\end{figure}

\begin{figure}[t]
  \centering
   \includegraphics[width=1\linewidth]{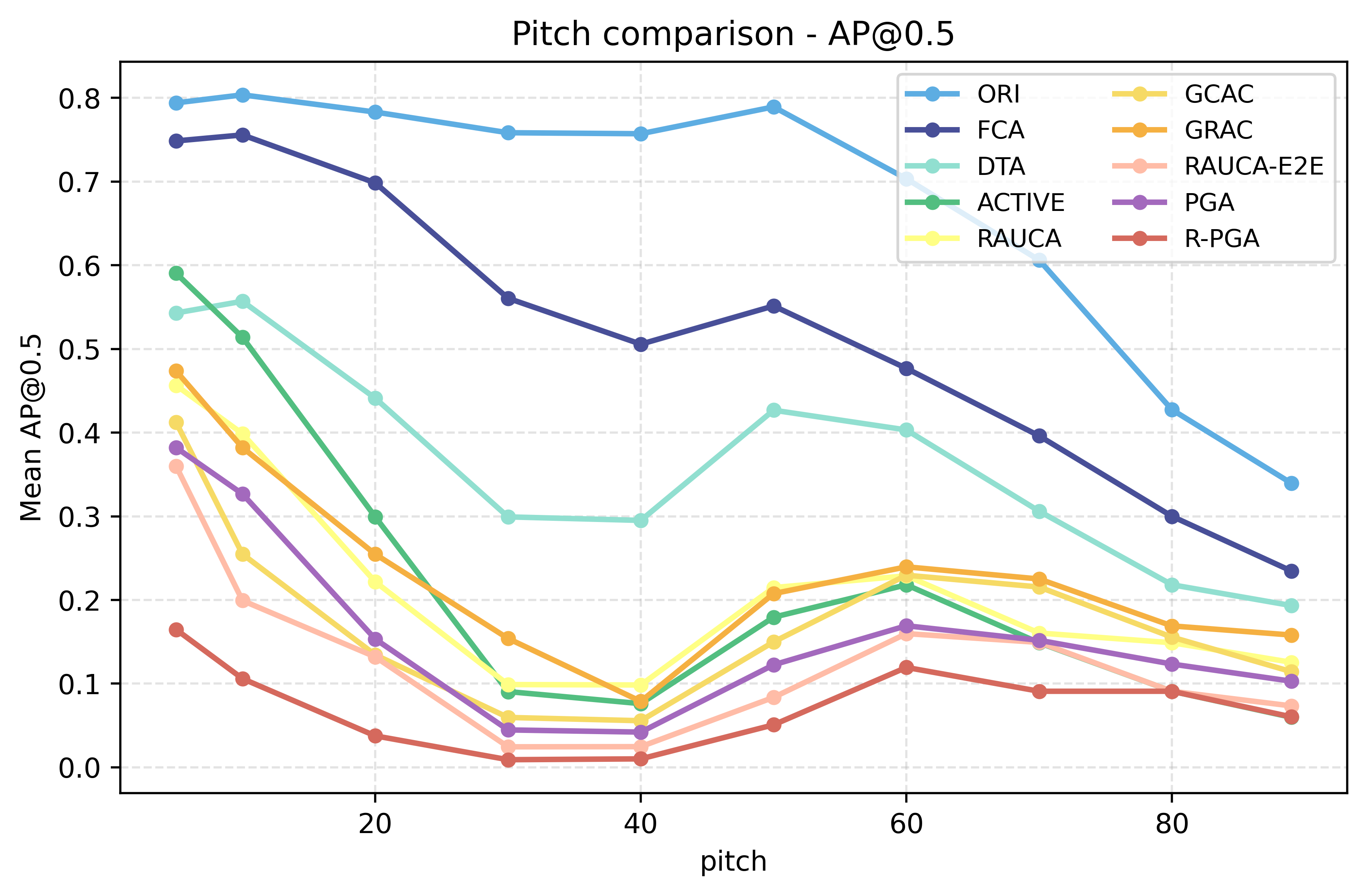}

   \caption{Comparison of detection results for different attack methods across various pitch angles, specifically reporting the average AP@0.5 for different detectors averaged over diverse weather conditions and distances.
}
   \label{fig:pitch}
   \vspace{-4mm}
\end{figure}

\begin{figure}[t]
  \centering
   \includegraphics[width=1\linewidth]{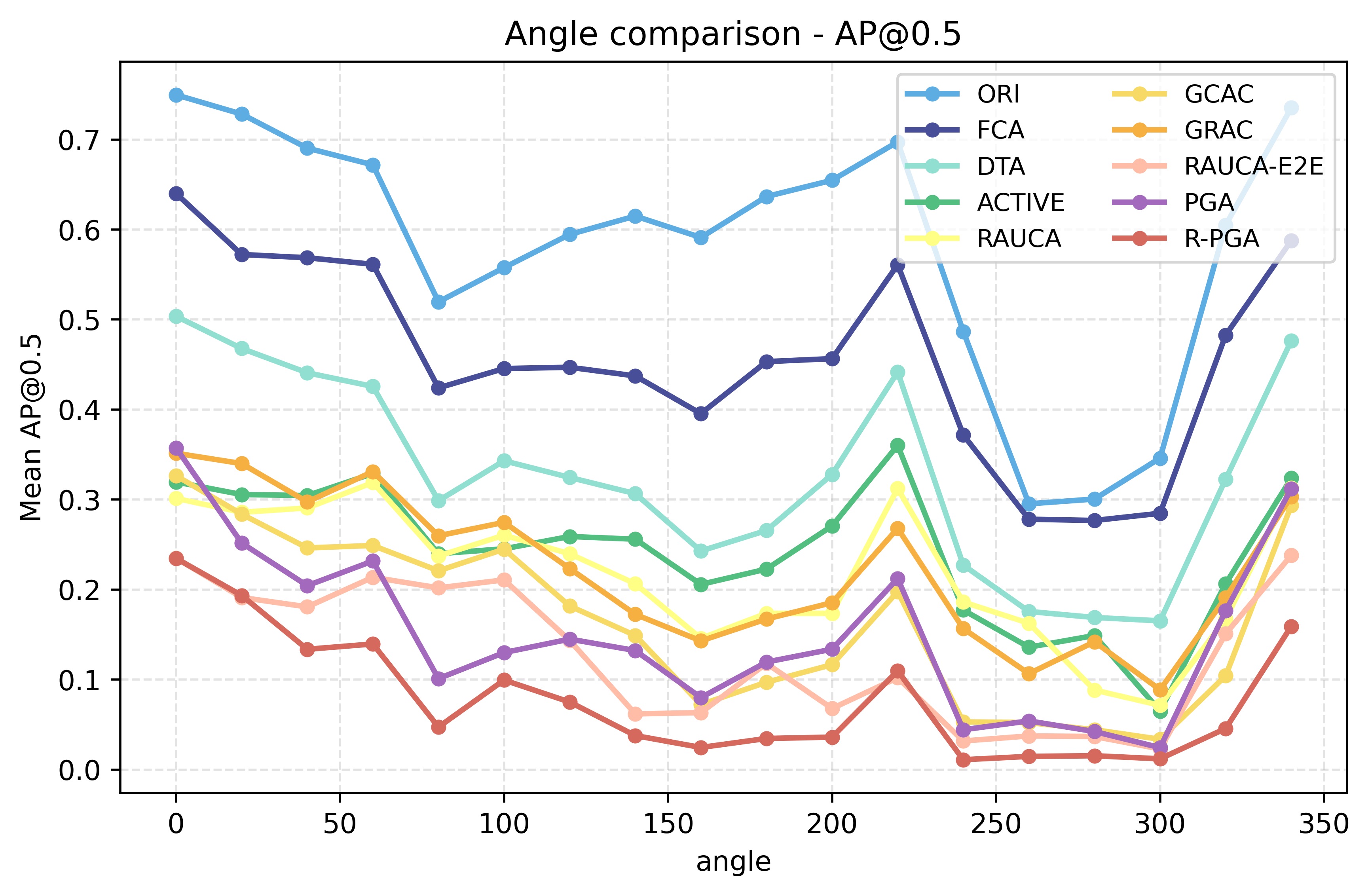}

   \caption{Comparison of detection results for different attack methods across various azimuth angles, specifically reporting the average AP@0.5 for different detectors averaged over diverse weather conditions and distances.
}
   \label{fig:angle}
   \vspace{-4mm}
\end{figure}

\begin{figure}[t]
  \centering
   \includegraphics[width=1\linewidth]{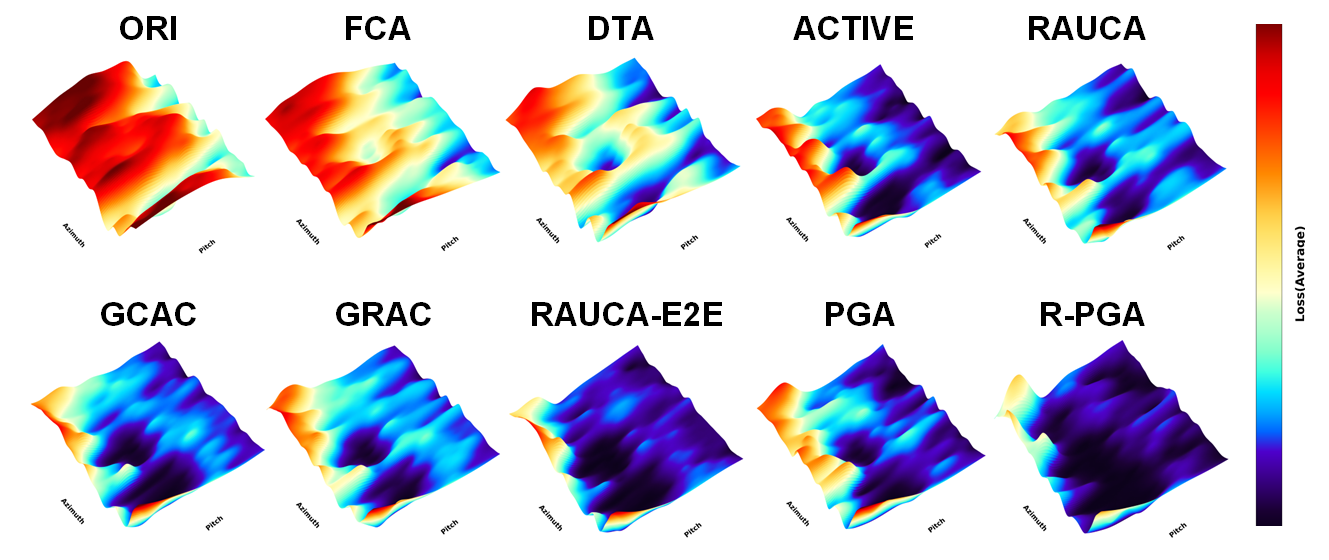}

   \caption{Visual comparison of loss landscapes for different attack methods across various physical configurations. For visualization clarity, we report the mean results averaged over weather conditions, distances, and target detectors.
}
   \label{fig:landscape}
\end{figure}

\section{Experiments}
In this section, we first detail the experimental settings and implementation details. We then validate the effectiveness of R-PGA through digital domain experiments, including extensive qualitative and quantitative comparisons and ablation studies. Finally, we present physical domain experiments, demonstrating the robust performance of the generated camouflage on a 1:24 scale toy car.

\begin{figure*}[t]
  \centering
   \includegraphics[width=1\linewidth]{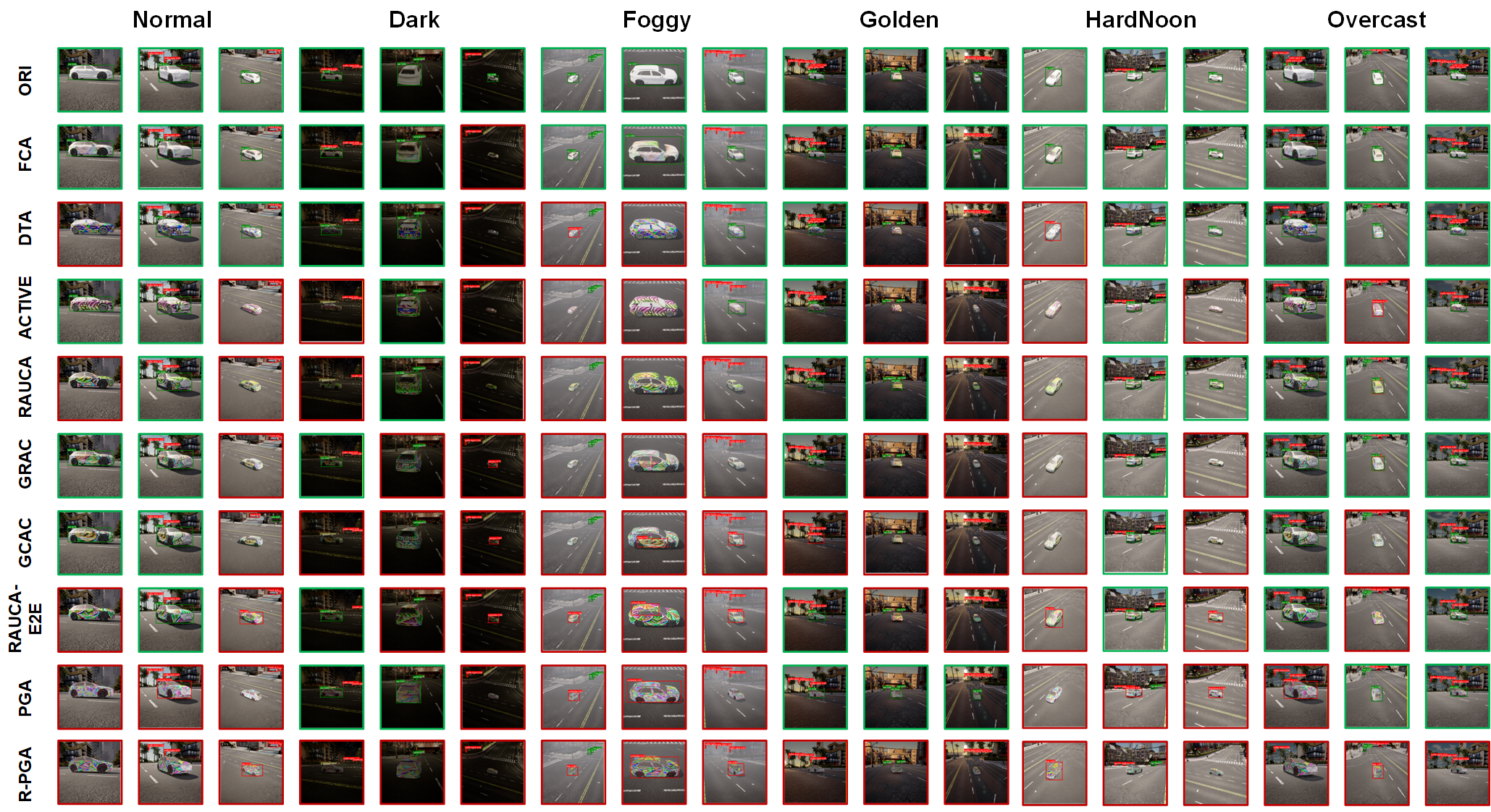}

   \caption{Visual comparison of detection results across different methods in the digital domain. We display camouflaged vehicles captured under diverse configurations, including varying pitch angles, azimuths, distances, and weather conditions. Green bounding boxes indicate correct detections (attack failure), while red bounding boxes denote incorrect detections (successful evasion).
}
   \label{fig:vis}
\end{figure*}

\begin{table*}[]
\caption{Comparison of AP@0.5 for different physical attack methods against vision foundation models. The reported results represent the average detection performance on the test dataset collected across multiple viewpoints, shooting distances, and weather conditions. Note that the adversarial camouflage is generated using Yolo-V3 and evaluated for black-box transferability (marked with *) on GLIP and DINO}
\centering
\rowcolors{1}{white}{gray!10}

\vspace{2mm}
\begin{tabular}{@{}c|cccccccccc@{}}
\toprule
                             & ORI    & FCA    & DTA    & ACTIVE & RAUCA  & RAUCA-E2E & GCAC   & GRAC   & PGA    & R-PGA           \\ \midrule
\multicolumn{1}{l|}{Yolo-V3} & 0.5251 & 0.4982 & 0.1934 & 0.3462 & 0.1869 & 0.1207    & 0.1515 & 0.1121 & 0.0318 & \textbf{0.0227} \\
DINO*                         & 0.8993 & 0.8903 & 0.8642 & 0.7872 & 0.8460 & 0.7674    & 0.7928 & 0.8778 & 0.8324 & \textbf{0.7421} \\
GLIP*                         & 0.9387 & 0.8732 & 0.7702 & 0.6970 & 0.7144 & 0.7037    & 0.6886 & 0.7589 & 0.7256 & \textbf{0.6749} \\ \bottomrule
\end{tabular}

\label{tab:foundation}
\end{table*}
\begin{table*}[t]
\centering
\caption{Ablation study of R-PGA components. The results verify the necessity of Physically Disentangled Reconstruction Module (Relit), Hybrid Rendering (HR), and the HPCM module for achieving robust attack performance (AP@0.5).}
\rowcolors{3}{white}{gray!10}

\vspace{2mm}
\scalebox{1.}{
\begin{tabular}{@{}ccc|cc|cc|cc|c@{}}
\toprule
\multirow{2}{*}{Relit} & \multirow{2}{*}{HR} & \multirow{2}{*}{HPCM} & \multicolumn{2}{c|}{One-Stage}    & \multicolumn{2}{c|}{Two-Stage}    & \multicolumn{2}{c|}{Transformer-based} & \multirow{2}{*}{Average} \\ \cmidrule(lr){4-9}
                            &                     &                       & Yolo-V3         & YoloX*          & FrRCN*          & MkRCN*          & D-DETR*            & PVT*              &                          \\ \midrule
                            & \checkmark                   & \checkmark                     & 0.0823          & 0.0556          & 0.0864          & 0.1025          & 0.0850             & 0.3445            & 0.1261                   \\
\checkmark                           &                     & \checkmark                    & 0.0681          & 0.0359          & 0.0479          & 0.1545          & 0.0469             & 0.2902            & 0.1073                   \\
\checkmark                           & \checkmark                   &                       & 0.0645          & 0.0344          & 0.0503          & 0.0866          & 0.0567             & 0.2881            & 0.0968                   \\
\checkmark                           & \checkmark                   & \checkmark                     & \textbf{0.0227} & \textbf{0.0325} & \textbf{0.0223} & \textbf{0.0565} & \textbf{0.0206}    & \textbf{0.2521}   & \textbf{0.0677}          \\ \bottomrule
\end{tabular}
}
\label{tab:ablation}
\end{table*}

\subsection{Experimental Setup}
\subsubsection{Datasets} 
To comprehensively validate the effectiveness of our attack method, we construct datasets for both digital domain and physical domain.

For the \textbf{digital domain dataset}, to enable a direct and fair comparison with prior methods utilizing the CARLA simulation environment~\cite{dosovitskiy2017carla}, we similarly construct our dataset using images collected from CARLA.
The test set for each attack method is generated by capturing images with a camera positioned around the vehicle deployed with the corresponding adversarial camouflage. 
We select 6 kinds of weather (dark, foggy, golden, hardnoon, normal, overcast), 4 distances ($5m, 10m, 15m, 20m$) and 10 camera pitch angles ($0^\circ, 10^\circ, 20^\circ, 30^\circ, 40^\circ, 50^\circ, 60^\circ, 70^\circ, 80^\circ, 90^\circ$).
For each setting, we conduct $360^\circ$ surrounding photography at $20^\circ$ intervals.
Consequently, for each comparative method, we construct a test set comprising 4,320 images.
For the \textbf{physical domain dataset}, we deploy the adversarial camouflages generated by R-PGA and other SOTA methods on a 1:24 scale Audi Q5 model car. 
The camouflage patterns are printed and applied using stickers. Subsequently, we capture images analogous to the digital domain dataset to construct a physical test set, enabling comprehensive qualitative and quantitative experiments across diverse scenarios. For detailed experimental settings, please refer to Sec.~\ref{sec:physet}.

\subsubsection{Target Models.}
We select 6 commonly used detection model architectures for the experiments, including one-stage detectors: Yolo-V3~\cite{redmon2018yolov3} and YoloX~\cite{ge2021yolox}; two-stage detectors: Faster R-CNN (FrRCN)~\cite{ren2015frcnn} and Mask R-CNN (MkRCN)~\cite{he2017maskrcnn}; transformer-based detectors: Deformable-DETR (D-DETR)~\cite{zhu2020ddetr} and PVT~\cite{wang2021pvt}, with all models pre-trained on the COCO dataset.

\subsubsection{Compared Methods.} 
We select 8 state-of-the-art physical adversarial attack methods as our baseline for comparison, including FCA (AAAI-2022)~\cite{wang2022FCA}, DTA (CVPR-2022)~\cite{suryanto2022DTA}, ACTIVE (ICCV-2023)~\cite{suryanto2023ACTIVE},  RAUCA (ICML-2024)~\cite{zhou2024rauca}, GCAC (IJCAI-2025)~\cite{liang2025gcac}, GRAC (ICCV-2025)~\cite{liang2025grac}, RAUCA-E2E (TDSC-2025)~\cite{zhou2025raucae2e}, PGA (ICCV-2025)~\cite{lou2025pga}.

\begin{figure*}[t]
  \centering
   \includegraphics[width=1\linewidth]{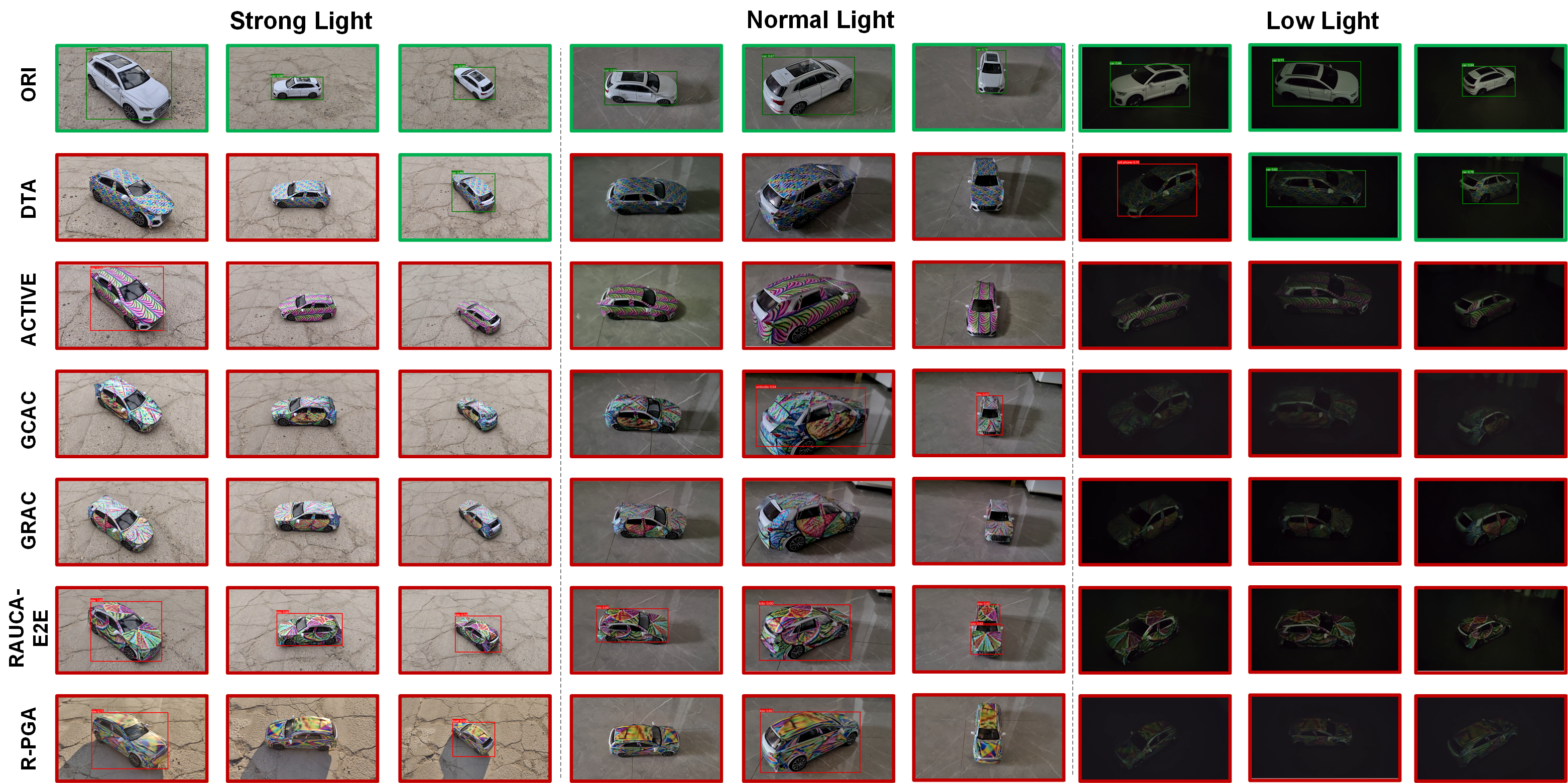}

   \caption{Visual comparison of detection results across different methods in the physical domain. We display camouflaged vehicles captured under diverse physical configurations, including varying pitch angles, azimuths, distances, and lighting conditions. Green bounding boxes indicate correct detections (attack failure), while red bounding boxes denote incorrect detections (successful evasion).
}
   \label{fig:phy}
\end{figure*}

\subsubsection{Evaluation Metrics.} 
To evaluate the effectiveness of various attack methods on detection models, we use AP@0.5($\%$), following~\cite{suryanto2022DTA, suryanto2023ACTIVE, zhou2024rauca, liang2025gcac, liang2025grac, zhou2025raucae2e, lou2025pga}, which is a standard measure capturing both recall and precision at a detection IoU threshold of 0.5.

\subsubsection{Training Details.}
We utilize the AdamW optimizer with a learning rate of 0.01 for R-PGA training and texture generation. 
We train the R-PGA framework for 20,000 iterations with a batch size of 8. For the HPCM module, we set the momentum parameter $\mu$ to 0.5 and the temperature parameter $\tau$ to 1. All experiments are conducted on a computing cluster equipped with four NVIDIA RTX 3090 (24GB) GPUs.

\subsection{Digital Experiments}
In this section, we present a comprehensive comparison of R-PGA and state-of-the-art methods, demonstrating the advantages of R-PGA. 
In these experiments, Yolo-V3 serves as the victim model for white-box attacks, with the adversarial camouflage transferred to five other detectors (marked with * throughout) to evaluate transferability.
\subsubsection{Digital World Attack}
We compare the average digital attack performance of R-PGA against state-of-the-art methods across various detectors and physical configurations.
Although R-PGA is capable of directly reconstructing and attacking using real-world photos, to ensure a fair comparison with other simulator-based methods, we capture clean vehicle images in CARLA to reconstruct the 3D scene and subsequently execute the R-PGA attack. 

First, we conducted a comprehensive comparative evaluation of physical adversarial effectiveness and transferability across diverse detection models, and the results are reported in Tab.~\ref{tab:main}.
R-PGA achieves state-of-the-art attack performance across all evaluated detectors.
Specifically, in the white-box setting (Yolo-V3), it drastically degrades the AP@0.5 to 0.0227.
Besides, R-PGA demonstrates robust transferability across diverse black-box architectures—including one-stage, two-stage, and transformer-based models—consistently outperforming prior arts like PGA and RAUCA-E2E. On average, our method reduces the detection AP to 0.0677, establishing a new benchmark for physical adversarial attacks.

Second, we conduct comparative experiments against state-of-the-art methods regarding weather conditions (Fig.~\ref{fig:weather}), shooting distances (Fig.~\ref{fig:distance}), pitch angles (Fig.~\ref{fig:pitch}), and azimuth angles (Fig.~\ref{fig:angle}). 
For each experimental setting, we report the average AP@0.5 computed over all variables excluding the controlled one, visualized via line charts and bar charts. Integrating these trends with the quantitative results in Tab.~\ref{tab:main}, R-PGA consistently establishes a new state-of-the-art. 
We attribute this superior adversarial effectiveness and cross-configuration robustness to the synergistic improvements in both simulation and optimization proposed in our framework: the High-Fidelity Relightable Scene Simulator bridges the domain gap to ensure radiometric stability against dynamic environmental lighting (Fig.~\ref{fig:weather}), while the Hard Physical Configuration Mining (HPCM) strategy addresses the optimization objective gap, actively flattening the loss landscape to guarantee geometric robustness across diverse viewing configurations (Figs.~\ref{fig:distance}-\ref{fig:angle}).

\subsubsection{Transferability to Vision Foundation Models}
We further evaluate the transferability of adversarial camouflages generated on YOLOv3 against two vision foundation models: GLIP~\cite{li2022glip} and DINO~\cite{zhang2022dino}. 
The corresponding AP@0.5 scores are reported in Tab.~\ref{tab:foundation}. 
Although these foundation models exhibit greater inherent robustness compared to conventional detectors, the results demonstrate that R-PGA consistently achieves the best average attack performance across all physical configurations. 
This empirically validates that R-PGA possesses superior adversarial transferability and robustness, effectively compromising even the most advanced large-scale vision systems.

\subsubsection{Visualization}

We present visualizations in Fig.~\ref{fig:vis} comparing the detection results of R-PGA with other attack methods in the digital domain. 
To fully demonstrate adversarial effectiveness and cross-configuration robustness, we selected a diverse set of configurations encompassing the various pitch angles, azimuths, weather conditions, and distances evaluated in our quantitative experiments. 
The results indicate that, compared to SOTA methods, R-PGA exhibits superior adversarial effectiveness across these varying physical configurations. 
Notably, even under particularly challenging configurations where adversarial camouflages from other methods fail (e.g., the comparison in the last column), R-PGA maintains effective attack performance.

\subsubsection{Comparison of Loss Landscapes}

We visualize the adversarial loss landscapes of different methods within the physical configuration space in Fig.~\ref{fig:landscape}, presented as 3D surface plots for clarity. 
Specifically, the two axes represent the azimuth and pitch angles, while the loss values are averaged across other configuration dimensions (i.e., weather conditions and shooting distances) as well as multiple detectors (consistent with the setting in Tab.~\ref{tab:main}, comprising one white-box and five black-box detectors). 
The results demonstrate that R-PGA, benefiting from the HPCM strategy and the high-fidelity hybrid rendering pipeline based on Relightable 3DGS, yields a loss landscape that is not only lower in average magnitude but also significantly flatter compared to other SOTA methods.
This empirically validates the superior adversarial robustness of our approach against configuration variations.

\subsubsection{Ablation Study}
We conduct an ablation experiment focusing on the three techniques in R-PGA, including Physically Disentangled Reconstruction Module (Relit), Hybrid Rendering Module (HR) and the Hard Physical Configuration Mining Module (HPCM), with the results shown in Tab.~\ref{tab:ablation}. It is apparent that using all three techniques simultaneously achieves the best attack performance.

\subsection{Physical Experiments}

\subsubsection{Experiment Settings}
\label{sec:physet}
We deploy adversarial camouflages generated by various SOTA methods and R-PGA on a 1:24 scale model car. To construct a comprehensive physical scene dataset, we capture images from multiple viewpoints—covering an azimuth range of $0^{\circ}$ to $360^{\circ}$ and pitch angles of $30^{\circ}$ and $60^{\circ}$—at distances ranging from 20 cm to 50 cm. Furthermore, we systematically incorporate three typical lighting conditions for data collection: Strong Light (SL), Normal Light (NL), and Low Light (LL). In total, we collect approximately 1,200 images per method for evaluation, employing YOLO-V3 as the victim detector.

\subsubsection{Evaluation Results}
\begin{table}[]
\rowcolors{1}{white}{gray!10}
\caption{Physical domain comparison of AP@0.5 against Yolo-V3 across different lighting conditions: Strong Light (SL), Normal Light (NL), and Low Light (LL), as well as the average. Each reported value represents the mean performance averaged over diverse pitch angles, azimuths, and shooting distances.}
\centering
\vspace{2mm}
\begin{tabular}{@{}c|ccc|c@{}}
\toprule
          & SL              & NL              & LL              & Average         \\ \midrule
ORI       & 0.9091          & 0.9134          & 0.9169          & 0.9131          \\
DTA       & 0.7872          & 0.8035          & 0.9074          & 0.8327          \\
ACTIVE    & 0.3466          & 0.2425          & 0.3958          & 0.3283          \\
GCAC      & 0.6354          & 0.2345          & 0.8151          & 0.5617          \\
GRAC      & 0.6144          & 0.4502          & 0.8163          & 0.6270          \\
RAUCA-E2E & 0.3611          & 0.2212          & 0.4534          & 0.3452          \\
R-PGA     & \textbf{0.3034} & \textbf{0.1849} & \textbf{0.3637} & \textbf{0.2840} \\ \bottomrule
\end{tabular}
\label{tab:phy}

\end{table}

Quantitative and qualitative results are reported in Tab.~\ref{tab:phy} and Fig.~\ref{fig:phy}, respectively. The results indicate that R-PGA drastically degrades the victim detector's performance from 0.9131 to 0.2840 (AP@0.5), representing a massive reduction of 0.6291. Furthermore, compared to the best-competing baseline RAUCA-E2E (0.3452), our method achieves an additional degradation of 0.0612. This validates that the high-fidelity reconstruction, hybrid rendering pipeline, and HPCM optimization strategy effectively guarantee the adversarial effectiveness and cross-configuration robustness of R-PGA in the physical domain.

\section{Conclusion}
In this paper, we introduce R-PGA, a novel framework for generating robust physical adversarial camouflage via Relightable 3D Gaussian Splatting. Our work identifies and addresses two fundamental limitations in prior physical attacks: the discrepancies in simulation fidelity (The Domain and Configuration Gap) and the pitfalls of average-case optimization (The Optimization Objective Gap).
To bridge the simulation gap, we propose a High-Fidelity Relightable Scene Simulator. By incorporating physically disentangled attributes into 3DGS and designing a hybrid rendering pipeline, we achieve photo-realistic scene reconstruction and precise lighting control, fundamentally resolving cross-view texture inconsistencies caused by entangled illumination. To close the optimization gap, we devise the Hard Physical Configuration Mining (HPCM) strategy. This approach actively mines and suppresses worst-case physical configurations, effectively flattening the rugged adversarial loss landscape to guarantee consistent robustness against geometric and radiometric variations.
Extensive experiments in both digital simulations and physical environments  demonstrate that R-PGA significantly outperforms state-of-the-art methods, establishing a new benchmark for physical adversarial attacks. By exposing the vulnerabilities of current detectors under complex dynamic scenarios, we hope this work serves as a solid foundation for future research on robust perception and defense in autonomous driving systems.

\ifCLASSOPTIONcaptionsoff
  \newpage
\fi



%
\bibliographystyle{IEEEtran}      
\footnotesize
\bibliography{egbib}

\section{Biography}

\vspace{-0.3in}
\begin{IEEEbiography}[{\includegraphics[height=0.9in,clip,keepaspectratio]{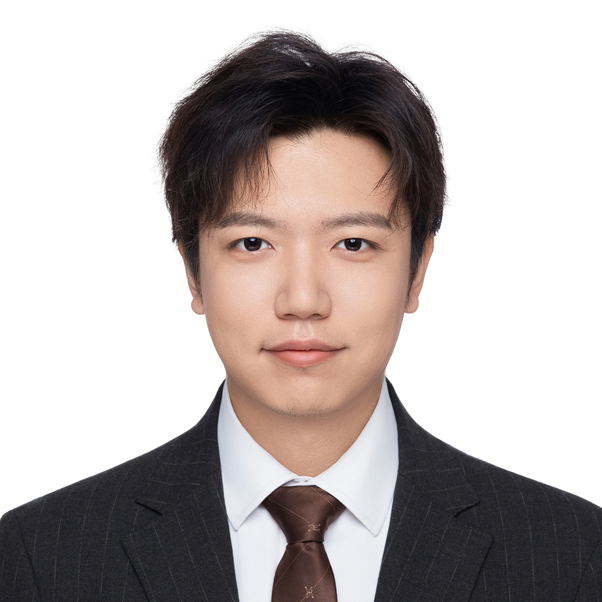}\hspace{0.3in}}]
{Tianrui Lou}
is currently pursuing the Ph.D. degree with the School of Cyber Science and Technology, Sun Yat-sen University, China. His research interests lie in trustworthy artificial intelligence and AI security, with a specific focus on physical adversarial attacks, 3D point cloud adversarial attacks, and adversarial training. He has authored several papers in top-tier conferences, including CVPR and ICCV.
\end{IEEEbiography}

\vspace{-0.3in}
\begin{IEEEbiography}[{\includegraphics[height=0.9in,clip,keepaspectratio]{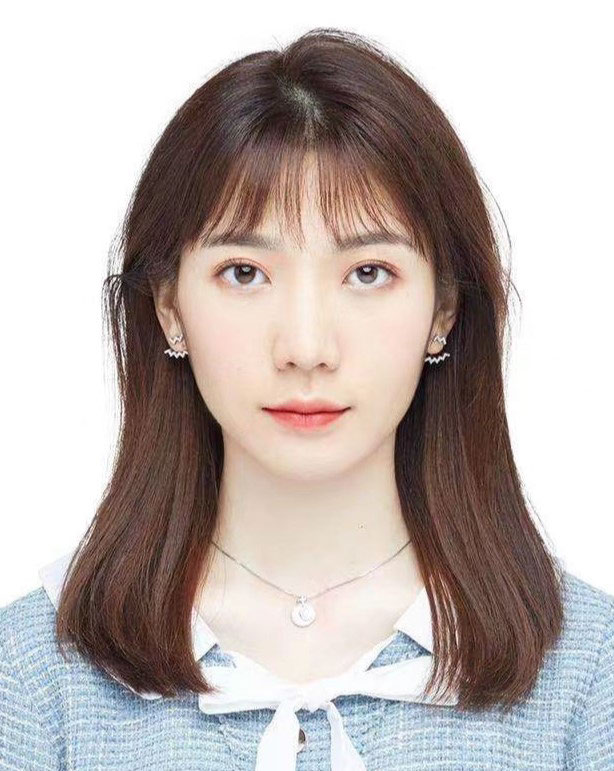}\hspace{0.3in}}]
{Siyuan Liang}
is currently a Research Fellow at the College of Computing \& Data Science at Nanyang Technological University. Her research interests span machine learning and computer vision, including trustworthy machine learning and security for deep object detection. In addition, she maintains a strong interest in the security of multimodal foundational models.
\end{IEEEbiography}

\vspace{-0.3in}
\begin{IEEEbiography}[{\includegraphics[height=0.9in,clip,keepaspectratio]{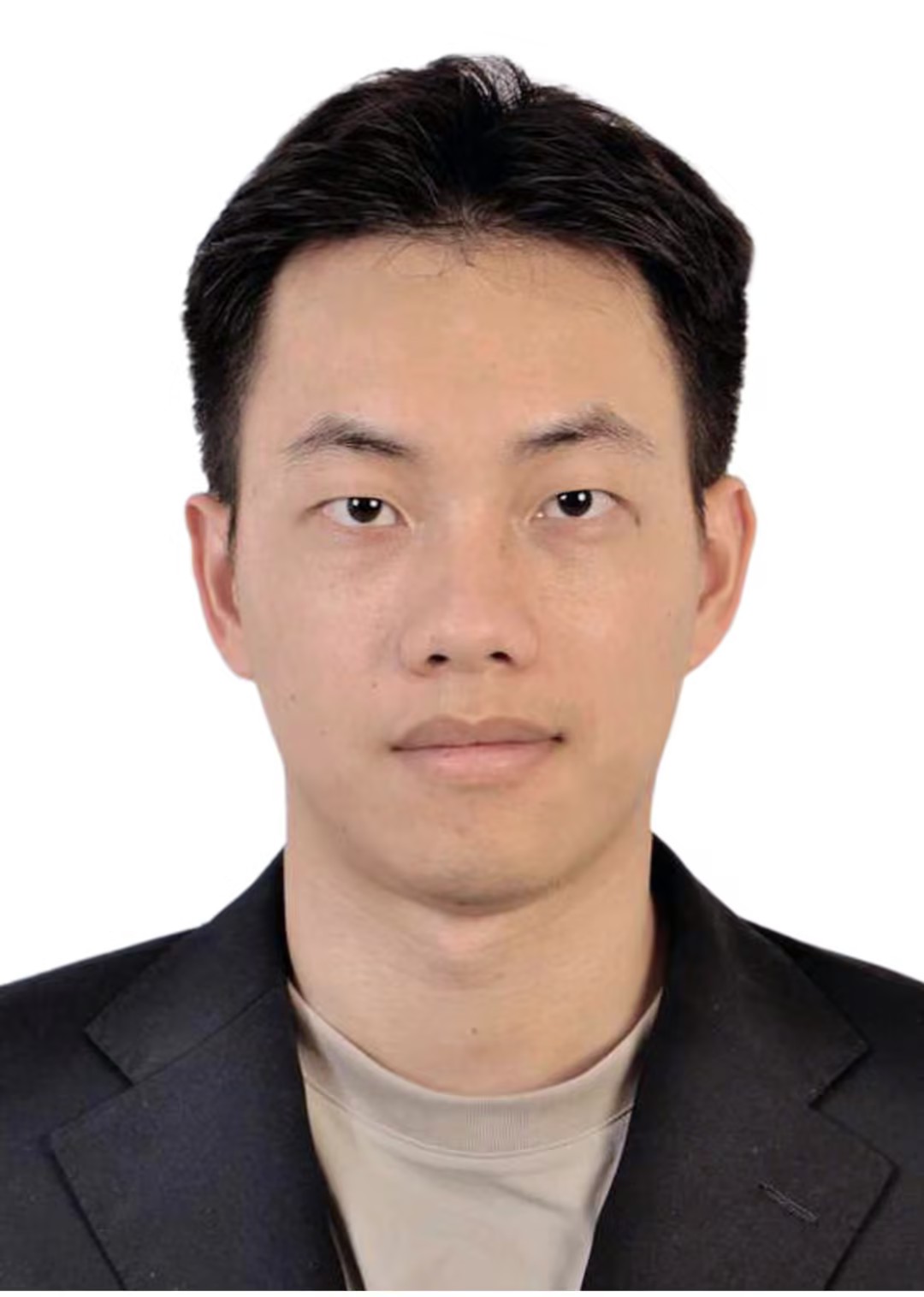}\hspace{0.3in}}]
{Jiawei Liang}
 is currently pursuing the Ph.D. degree with the School of Cyberscience and Technology, Sun Yat-sen University. His research interests include adversarial attacks and backdoor learning in computer vision models. He has authored several papers in top-tier conferences and journals, including ICLR, ICCV, and IJCV.
\end{IEEEbiography}

\vspace{-0.3in}
\begin{IEEEbiography}[{\includegraphics[height=0.9in,clip,keepaspectratio]{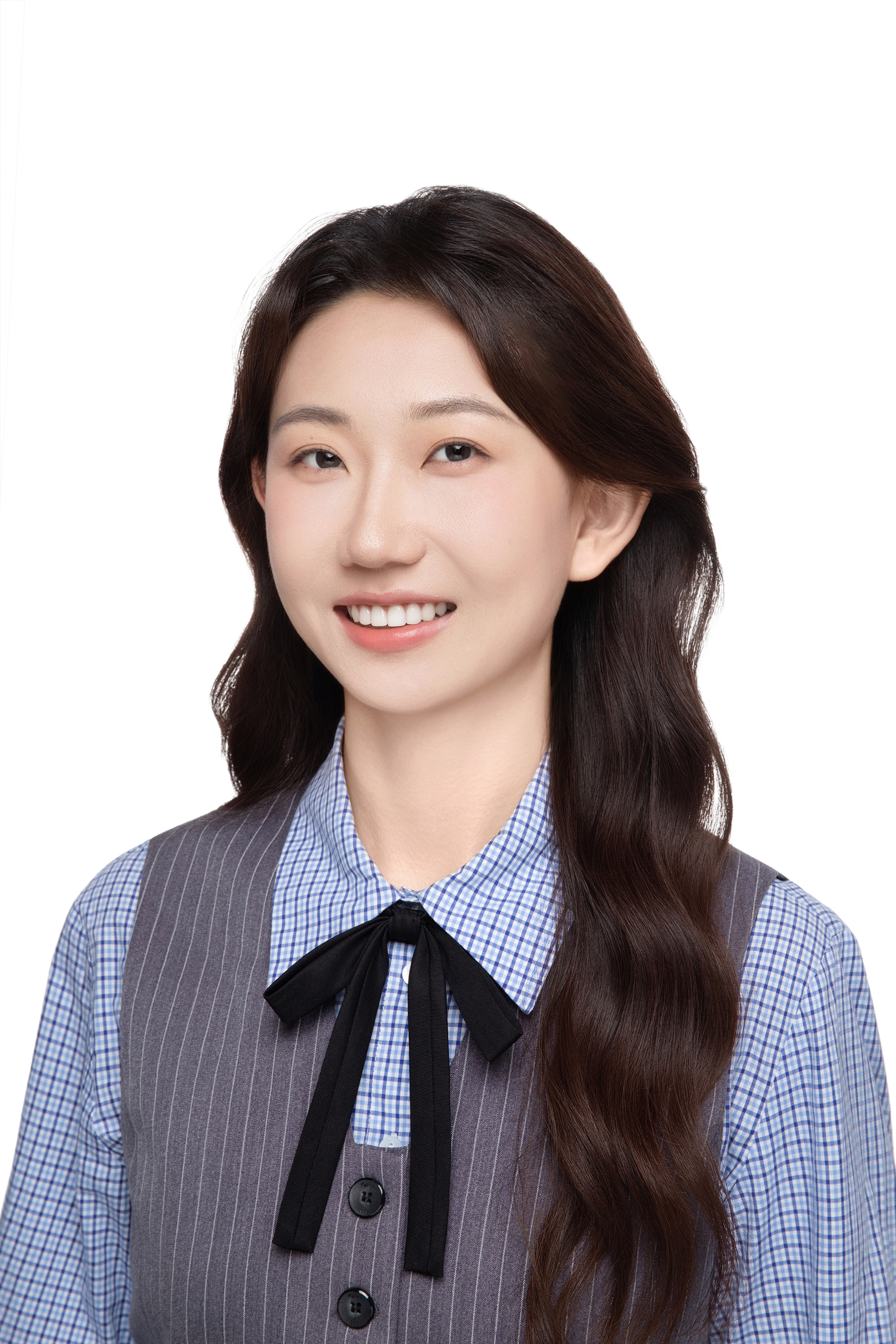}\hspace{0.3in}}]
{Yuze Gao}
is currently pursuing the Ph.D. degree with the School of Intelligent Systems Engineering, Sun Yat-sen University, China. Her research interests lie in trustworthy artificial intelligence and AI security, with a specific focus on enhancing the transferability of adversarial examples and 3D point cloud adversarial attacks. Additionally, she is also interested in 3D reconstruction.
\end{IEEEbiography}

\vspace{-0.3in}
\begin{IEEEbiography}[{\includegraphics[height=0.9in,clip,keepaspectratio]{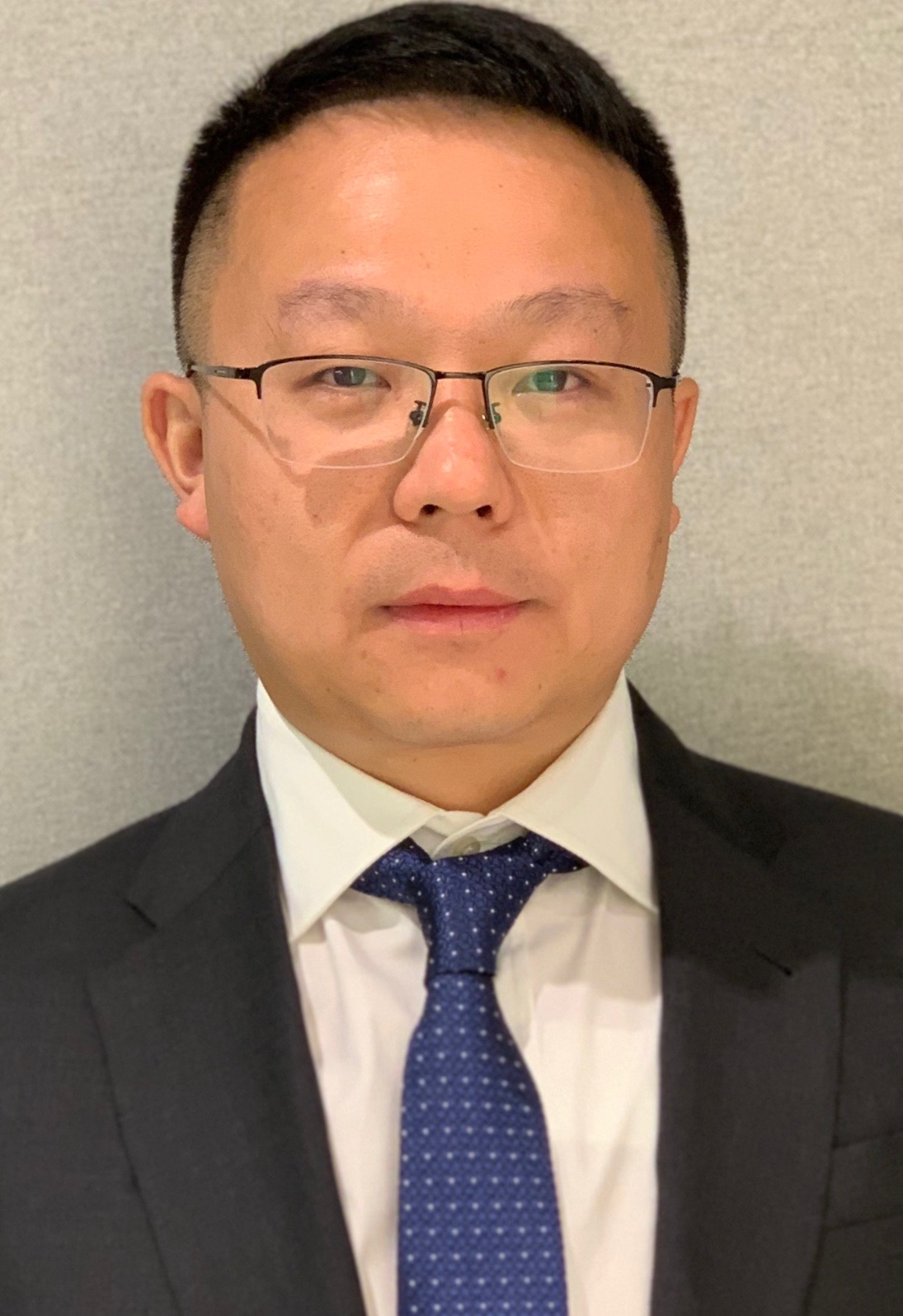}\hspace{0.3in}}]
{Dr. Xiaochun Cao}
 received the B.S. and M.S. degrees in computer science from Beihang University, China, and the Ph.D. degree in computer science from the University of Central Florida, USA. He is with the School of Cyber Science and Technology, Sun Yat-sen University, China, as a Full Professor and the Dean. He has authored and co-authored multiple top-tier journal and conference papers. He is on the Editorial Boards of the IEEE TIP, IEEE TMM, IEEE TCSVT. He was the recipient of the Best Student Paper Award at the ICPR (2004, 2010). He is the Fellow of the IET.
\end{IEEEbiography}
\clearpage

\appendices

\section{Image Translation Model}
We construct a training dataset using the CARLA simulator, comprising multi-view images of identical scenes captured under diverse lighting and weather conditions. 
Formally, we define a training tuple as 
\begin{equation}
    \mathcal{U} = \{ (\bm{I}_s, \bm{E}_s), (\bm{I}_t, \bm{E}_t) \}, 
\end{equation}
where $\bm{I}$ and $\bm{E}$ represent the captured scene image and the corresponding environment map, respectively, with subscripts $s$ and $t$ denoting the source and target domains.
Utilizing the Segment Anything Model (SAM) \cite{kirillov2023sam}, we disentangle the foreground ($f$) and background ($b$) components, yielding the set $\{\bm{I}_s^f, \bm{I}_s^b, \bm{I}_t^f, \bm{I}_t^b\}$.
Specifically, our training dataset is collected using the CARLA simulator, comprising 18 distinct vehicle models (e.g., Audi, Ford, Mini, Tesla, etc.). The dataset covers five weather conditions (Dusk, HarshSun, Night, Overcast, and Sunny), with shooting distances ranging from 5m to 10m. The viewing configurations include pitch angles from $0^{\circ}$ to $90^{\circ}$ with a step of $10^{\circ}$, and azimuth angles from $0^{\circ}$ to $360^{\circ}$ with a step of $60^{\circ}$. By pairing distinct weather conditions as source and target domains, we collected a total of 21,600 image pairs.

Our goal is to train a conditional generative model to synthesize the target background $\bm{I}_t^b$. 
To ensure the generated background is photometrically consistent with the relighted foreground and the new environment, we introduce a composite conditioning vector $\mathbf{v}$.
This vector encodes the target environment map $\bm{E}_t$, the target relighted foreground $\bm{I}_t^f$, and the intrinsic appearance contrast of the source domain, formulated as:
\begin{equation}
    \mathbf{v} = \mathcal{E}(I_s^b) \oplus \mathcal{E}(I_t^f - I_s^f) \oplus \psi(E_t),
\end{equation}
where $\mathcal{E}(\cdot)$ denotes the encoder transforming images into the latent space, $\psi(\cdot)$ is an embedding function for the environment map, and $\oplus$ represents the concatenation operation.
The training objective is to optimize a velocity field $v_\theta$ that transports the source distribution to the target background distribution. We employ the flow matching objective defined as:
\begin{equation}
    \mathcal{L} = \mathbb{E}_{t, z_0, z_1} \left[ \left\| v_t - v_\theta(z_t, t, \mathbf{c}) \right\|^2 \right],
\end{equation}
where $z_1 = \mathcal{E}(I_t^b)$ represents the latent representation of the ground truth target background, $z_t$ is the intermediate state at timestep $t$, and $v_t$ is the target velocity field connecting the noise distribution to $z_1$. Through this formulation, the model learns to effectively hallucinate the target background $I_t^b$ in a single inference step.

\section{HPCM Optimization Objective} 

In this section, we provide the detailed mathematical derivation to prove that the Hard Physical Configuration Mining (HPCM) strategy is equivalent to optimizing the Log-Sum-Exp (LSE) objective, thereby minimizing the worst-case physical adversarial loss.

\textbf{1. Problem Setup}
Let $\mathcal{L}_i(\mathbf{a}) = \mathcal{L}(\mathbf{a}, \bm{c}_i)$ denote the adversarial detection loss given the albedo $\mathbf{a}$ under the $i$-th physical configuration $\bm{c}_i$, where $i \in \{1, 2,\dots,q\}$.
The HPCM module samples configurations based on a probability distribution $P(c_i)$ defined by the Softmax of the difficulty scores:

\begin{equation}
    P(c_i) = \frac{\exp(\mathcal{L}_i/\tau)}{\sum_{j=1}^{q} \exp(\mathcal{L}_j/\tau)},
\end{equation}
where $\tau$ is the temperature parameter.
During the iterative optimization, the update direction for albedo $a$ is determined by the expected gradient under this distribution:
\begin{equation}
    \mathbf{g}_{\text{HPCM}} = \mathbb{E}_{\bm{c} \sim P} [\nabla_{\mathbf{a}} \mathcal{L}(\bm{c})] = \sum_{i=1}^{q} P(\bm{c}_i) \cdot \nabla_{\mathbf{a}} \mathcal{L}_i(\mathbf{a}).
\end{equation}

\textbf{2. Gradient of the Log-Sum-Exp Objective}
We define the global robust objective function using the Log-Sum-Exp (LSE) function:
\begin{equation}
    \mathcal{J}_{\text{LSE}}(\mathbf{a}) = \tau \log \left( \sum_{j=1}^{M} \exp \left( \frac{\mathcal{L}_j(\mathbf{a})}{\tau} \right) \right).
\end{equation}

To find the optimization direction for $\mathcal{J}_{\text{LSE}}$, we compute its gradient with respect to the albedo parameters $\mathbf{a}$ using the chain rule:

\begin{equation}
\begin{aligned}
\nabla_{\mathbf{a}} \mathcal{J}_{\text{LSE}} &= \nabla_{\mathbf{a}} \left[ \tau \log \left( \sum_{j=1}^{M} \exp \left( \frac{\mathcal{L}_j(\mathbf{a})}{\tau} \right) \right) \right] \\
&= \tau \cdot \frac{1}{\sum_{j=1}^{M} \exp \left( \frac{\mathcal{L}_j(\mathbf{a})}{\tau} \right)} \cdot \nabla_{\mathbf{a}} \left( \sum_{k=1}^{M} \exp \left( \frac{\mathcal{L}_k(\mathbf{a})}{\tau} \right) \right).
\end{aligned}
\end{equation}

Next, we compute the gradient of the summation term:

\begin{equation}
\begin{aligned}
\nabla_{\mathbf{a}} \left( \sum_{k=1}^{M} \exp \left( \frac{\mathcal{L}_k(\mathbf{a})}{\tau} \right) \right) &= \sum_{k=1}^{M} \exp \left( \frac{\mathcal{L}_k(\mathbf{a})}{\tau} \right) \cdot \nabla_{\mathbf{a}} \left( \frac{\mathcal{L}_k(\mathbf{a})}{\tau} \right) \\
&= \sum_{k=1}^{M} \exp \left( \frac{\mathcal{L}_k(\mathbf{a})}{\tau} \right) \cdot \frac{1}{\tau} \cdot \nabla_{\mathbf{a}} \mathcal{L}_k(\mathbf{a}).
\end{aligned}
\end{equation}

Substituting this back into the expression for $\nabla_{\mathbf{a}} \mathcal{J}_{\text{LSE}}$, the constant $\tau$ cancels out:

\begin{equation}
\begin{aligned}
\nabla_{\mathbf{a}} \mathcal{J}_{\text{LSE}} &= \frac{1}{\sum_{j=1}^{M} \exp \left( \frac{\mathcal{L}_j}{\tau} \right)} \cdot \sum_{k=1}^{M} \exp \left( \frac{\mathcal{L}_k}{\tau} \right) \nabla_{\mathbf{a}} \mathcal{L}_k \\
&= \sum_{k=1}^{M} \left( \frac{\exp(\mathcal{L}_k/\tau)}{\sum_{j=1}^{M} \exp(\mathcal{L}_j/\tau)} \right) \cdot \nabla_{\mathbf{a}} \mathcal{L}_k.
\end{aligned}
\end{equation}

\textbf{3. Equivalence and Bounds Analysis}
Comparing the term in the parentheses with Eq. (A.1), we observe that it is identical to the sampling probability $P(c_k)$. Thus:
\begin{equation}
    \nabla_{\mathbf{a}} \mathcal{J}_{\text{LSE}} = \sum_{k=1}^{M} P(c_k) \cdot \nabla_{\mathbf{a}} \mathcal{L}_k = \mathbf{g}_{\text{HPCM}}.
\end{equation}

\textbf{Conclusion: }This equality proves that applying HPCM is mathematically equivalent to performing gradient descent on the $\mathcal{J}_{LSE}$ objective.
Furthermore, according to convex analysis theory, the LSE function is bounded by the maximum function:
\begin{equation}
    \max_{i} \mathcal{L}_i \le \mathcal{J}_{\text{LSE}}(\mathbf{a}) \le \max_{i} \mathcal{L}_i + \tau \log M.
\end{equation}

This inequality indicates that minimizing $\mathcal{J}_{LSE}$ effectively minimizes the upper bound of the worst-case configuration. 
By suppressing the maximum loss $\max \mathcal{L}_i$, the optimization process actively flattens the peaks in the loss landscape, thereby enhancing the overall robustness of the physical camouflage.

\end{document}